\pdfoutput=1
\PassOptionsToPackage{dvipsnames, svgnames}{xcolor}
\documentclass[11pt]{article}
\usepackage{xcolor}
\definecolor{c1}{HTML}{488f31}
\definecolor{c2}{HTML}{6e9b3c}
\definecolor{c3}{HTML}{90a84a}
\definecolor{c4}{HTML}{afb45c}
\definecolor{c5}{HTML}{ccc170}
\definecolor{c6}{HTML}{e6ce87}
\definecolor{c7}{HTML}{ffdba0}
\definecolor{c8}{HTML}{f9c488}
\definecolor{c9}{HTML}{f4ac74}
\definecolor{c10}{HTML}{ee9264}
\definecolor{c11}{HTML}{e77859}
\definecolor{c12}{HTML}{df5c53}

\definecolor{p1}{HTML}{bbc5a4}
\definecolor{p2}{HTML}{ede8d9}
\definecolor{p3}{HTML}{e0b992}

\usepackage{ACL2023}
\usepackage{amsmath}
\usepackage{graphicx}
\usepackage{tabularx}
\usepackage{makecell}
\usepackage{caption}
\usepackage{color, colortbl, booktabs}
\usepackage{multirow}
\usepackage{subcaption}
\usepackage{rotating}
\usepackage{tikz}
\usepackage{booktabs}
\usepackage{times}
\usepackage{makecell}
\usepackage{latexsym}
\usepackage[T1]{fontenc}
\usepackage{CJKutf8}
\usepackage[export]{adjustbox}[2011/08/13]

\usepackage[utf8]{inputenc}

\usepackage{microtype}
\usepackage{multirow}
\usepackage{pifont}

\usepackage{inconsolata}

\title{\textit{Let LLMs Take on the Latest Challenges !} \\ A Chinese Dynamic Question Answering Benchmark}

\author{Zhikun Xu\thanks{\ \ indicates equal contribution.},
  Yinghui Li$^*$,
  Ruixue Ding\thanks{\ \ Corresponding authors.},
  Xinyu Wang,
  Boli Chen, 
  Yong Jiang$^\dagger$ \\
  \textbf{Hai-Tao Zheng,  
  Wenlian Lu,
  Pengjun Xie,  
  Fei Huang}\\
  Institute for Intelligent Computing, Alibaba Group \\
  \texttt{\{ada.drx, yongjiang.jy\}@alibaba-inc.com} \\
  \texttt{Dataset \& Code:} \url{https://github.com/Alibaba-NLP/CDQA}
  }

\begin{document}
\maketitle

\begin{abstract}
How to better evaluate the capabilities of Large Language Models (LLMs) is the focal point and hot topic in current LLMs research.
Previous work has noted that due to the extremely high cost of iterative updates of LLMs, they are often unable to answer the latest dynamic questions well.
To promote the improvement of Chinese LLMs' ability to answer dynamic questions, in this paper, we introduce \textbf{CDQA}, a \textbf{C}hinese \textbf{D}ynamic \textbf{QA} benchmark containing question-answer pairs related to the latest news on the Chinese Internet. We obtain high-quality data through a pipeline that combines humans and models, and carefully classify the samples according to the frequency of answer changes to facilitate a more fine-grained observation of LLMs' capabilities.
We have also evaluated and analyzed mainstream and advanced Chinese LLMs on \textit{CDQA}. Extensive experiments and valuable insights suggest that our proposed \textit{CDQA} is challenging and worthy of more further study.
We believe that the benchmark we provide will become one of the key data resources for improving LLMs' Chinese question-answering ability in the future.
\end{abstract}

\section{Introduction}
\begin{CJK*}{UTF8}{gbsn}
Due to the excellent emergence capabilities and unified task paradigm, Large Language Models (LLMs) are undoubtedly the more popular stars in the field of Natural Language Processing (NLP) or Artificial Intelligence~\cite{wei2022emergent, DBLP:journals/corr/abs-2307-09007, shanahan2024talking}.
To promote the improvement of LLMs capabilities, more and more researchers have invested in building various LLMs evaluation benchmarks~\cite{chang2023survey, huang2023lateval}. In the era of LLMs, high-quality evaluation benchmarks allow researchers to better understand the capabilities of LLMs, thereby stimulating further research on how to enhance LLMs.

Question answering is an important and long-standing topic in NLP~\cite{rajpurkar2016squad, joshi2017triviaqa, he2018dureader}. 
Especially for LLMs, QA tasks have almost become the indispensable basic task in LLMs research~\cite{pan2024unifying}.
Various forms of QA benchmarks can be used to measure the capabilities of LLMs in different dimensions~\cite{ adlakha2022topiocqa, csrr-2022-commonsense, rein2023gpqa,  huang2023ceval}.
Recently, the introduction of English FreshQA~\cite{vu2023freshllms} has attracted widespread attention. It challenges LLMs through questions with dynamically changing answers, aiming to test LLMs' mastery of the latest factual knowledge. Obviously, being able to answer the latest questions determines to some extent whether LLMs can truly move towards large-scale daily applications.
\textbf{Urgently, we note that there is still no such benchmark in the Chinese community, although LLMs in the Chinese scenario still face the same challenges and dilemmas}, as shown in Table~\ref{tab:intro}.

\begin{table}[]
\small
\centering
\begin{tabular}{l|lc}
\toprule
\midrule
\textbf{Static} & ACL 主会每年举办几次？ & \\
\textbf{Question} & How many times does the ACL & \\
 & annual meeting take place each year? & \\
\midrule
\textbf{GPT-4's} & 一年一次。 & \multirow{2}{*}{\ding{51}} \\
\textbf{Answer} & Once a year. &\\ 
\midrule
\textbf{Dynamic} & 下一次 ACL 将在哪里举办？ & \\
\textbf{Question} & Where will the next ACL be held? & \\
\midrule
\textbf{GPT-4's} & 我无法提供相关信息。 & \multirow{2}{*}{\ding{55}} \\
\textbf{Answer} & I can't provide the information. & \\ 
\midrule
\bottomrule
\end{tabular}
\caption{Examples of static and dynamic questions. The \textbf{GPT-4} is on Feb 11, 2024.}
\label{tab:intro}
\end{table}

To let LLMs in Chinese scenarios take on the latest challenges and empower them to answer dynamic questions, in this work, we present \textbf{CDQA}, a \textbf{C}hinese \textbf{D}ynamic \textbf{QA} benchmark. 
Specifically, we design a semi-automatic data production pipeline to construct our benchmark. In this pipeline, we first automatically generate a large number of raw queries with the help of two LLMs with different roles, one is to extract key entities from the latest Chinese news, and the other is to automatically generate question queries based on the extracted entities that will be as the corresponding answers. 
Then we ask the well-trained annotators to filter, rewrite, and classify the automatically generated question samples to ensure the quality of \textit{CDQA}. 
Through such a semi-automatic data construction method with human participation, we obtain 1,339 question-answer pairs for \textit{CDQA}, classified by how frequently their answers change (i.e., fast-changing, slow-changing, and never-changing). The purpose of classifying \textit{CDQA} samples by the frequency of answer changes is to provide finer-grained evaluation for LLMs, facilitating researchers to better perceive the true performance of LLMs.

Based on our constructed \textit{CDQA}, we select a series of widely used and advanced LLMs in the Chinese community for evaluation. Results show that \textbf{GPT-4} still ranks at the top with searched results from search engines, surpassing at most \textbf{nearly 10 F1-recall scores} with less hallucination than the second-best model, i.e., \textbf{Deepseek-67B-Chat}, although Chinese-oriented LLMs tend to have more internal knowledge than OpenAI models. Besides, \textbf{in-context learning} and \textbf{prompting methods} like Chain-of-Thought generally increase performances with searched evidence but also elicit more hallucinations in LLMs. For \textbf{search engines} in the open-book scenario, Google consistently takes advantage over Bing for all baseline models, showing great strength as a retriever for LLMs.

In summary, our contributions could be summarized as follows:
\begin{enumerate}
    \item We first introduce the idea of using dynamic questions to challenge Chinese LLMs, which provides a new direction for the development of LLMs in Chinese community.
    \item We construct the high-quality \textit{CDQA} benchmark composed of dynamic questions, which will become an important data resource for promoting the progress of Chinese LLMs.
    \item Extensive experiments and detailed analyses based on \textit{CDQA} provide valuable insights and discoveries, which are instructive for subsequent research about how to enhance LLMs to handle dynamic questions.
\end{enumerate}

\section{Chinese Dynamic Question Answering (CDQA)}
\subsection{Overview}
\begin{figure*}
  \center{
  \vspace{-5mm}
  \includegraphics
  [width=\linewidth]
  {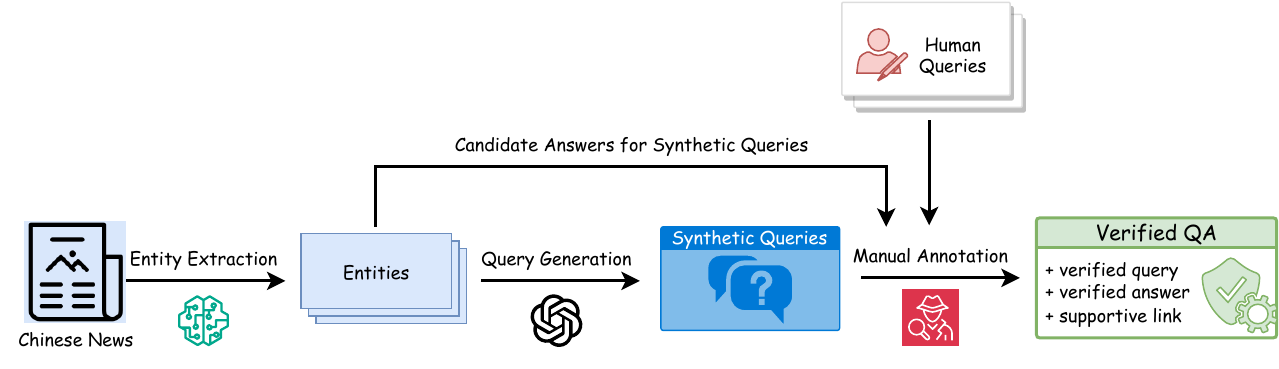}}
  \caption{Data Generation Pipeline for \textbf{CDQA} dataset. We first collect Chinese News from Internet and then extract entities from these news passages. Based on GPT-4, we generate synthetic queries from passages and corresponding entities. Manual annotation is conducted to verify the synthetic data and extra human-crafted queries, providing the verified queries, answers and supportive evidence links.}
  \label{main_pipeline}
\end{figure*}

Our dataset, \textbf{CDQA}, mainly originates from latest news in Chinese Internet from different areas such as finance, daily life, politics, technology and so on. Besides, there are also queries collected from Chinese labelers. They represent the information-seeking cases of Chinese people. The generation pipeline could be illustrated in Figure~\ref{main_pipeline}. The dataset currently consists of 1,339 questions covering a range of topics with evolving answers which are mostly extracted entities from the raw corpus scraped from Chinese Internet and it is being regularly updated.

\subsection{Data Collection}
We collect \textit{CDQA} dataset in two stage. \textbf{The first stage is automatic generations with Entity Extraction and Doc2Query}, for which we use SeqGPT \cite{yu2023seqgpt}, and GPT-4 \cite{openai2023gpt4}, which could give great amount of raw question-answering pairs as SeqGPT extracts entities from latest Chinese news and GPT-4 is prompted into generating corresponding questions. For GPT-4 prompts, we use few-shot prompting in generating diverse questions from entities. \textbf{The second stage is manual labeling from crowd-sourced workers}. The Chinese labelers not only filter questions which are answered with biases, ambiguities and obsolete\footnote{The answer should be only supported with the knowledge after January 1st 2019 except for static knowledge, i.e., never-changing.} knowledge but also annotate with \textbf{tags}, check the correctness and \textbf{rewrite} the question answer pairs to be more time-related and dynamic. At the very beginning, the labelers are shown with pre-annotation examples and annotation guides.

\paragraph{Tags} The tags are annotated for questions and answers. For questions, we have the same taxonomy as \textit{FreshQA} \cite{vu2023freshllms}. The questions are categorized as \textbf{fast-changing}, \textbf{slow-changing}, and \textbf{never-changing}. For answers, we categorize these entities or short texts as \textbf{person}, \textbf{location}, \textbf{time}, \textbf{event}, \textbf{artificial work}, \textbf{group}, \textbf{nature}, \textbf{quantity} and \textbf{other}. Therefore, we could evaluate the models' latest world knowledge from various perspectives. The taxonomy is illustrated in Appendix~\ref{sec:tag_taxonomy}.

\paragraph{Quality Control} After getting the synthetic queries, the human annotators could rewrite and calibrate the questions and answers to make the QA pairs correct, consistent and dynamic. For example, annotators are required to provide the supporting evidence \texttt{URLs} along with correct answers using search engines. This calibration process could solidify our answers with supplementary valid information and help us better iterate the dataset as the generation process in the previous stage is not well-evaluated with supportive documents, let alone the correctness. Moreover, in order to facilitate the periodic updates, we filter out the questions with more than one valid answer.

For inter-annotator agreement, we randomly sample 100 examples from synthetic question-answer pairs and annotations from two annotators in the same annotation vendor are measured by \textbf{acceptance} (\textit{whether the pair is accepted or discarded}), \textbf{question tags} and \textbf{answer types}. The ground-truth labels are provided by authors. For each category, we calculate their Cohen Kappa scores~\cite{cohenkappa}. From Table~\ref{tab:annotator-agreement}, the averaged score across all types of annotations are above 63.1, representing ``substantial agreement'' for our dataset annotations.
\begin{table}
\setlength\tabcolsep{2pt}
\small
\centering
\begin{tabular}{lccc}
\toprule
 & \textbf{Acceptance} & \textbf{Question Tags} & \textbf{Answer Types} \\ 
 \cmidrule{2-4}
Ann1 v.s. Ann2 & 62.3 & 87.2 & 96.6\\
GT v.s. Ann1 & 79.6 & 59.1 & 100\\
GT v.s. Ann2 & 47.3 & 68.3 & 100\\
\cmidrule{2-4}
Avg & \textbf{63.1} & \textbf{71.5} & \textbf{98.9} \\
\bottomrule
\end{tabular}%
\caption{Inter-annotator agreement for different annotation sections are calculated by \textbf{Cohen Kappa scores}. Ann1/2 represents Annotator1/2 respectively and GT represents Ground Truth. Our annotations could be considered as ``substantial agreement'' as the average scores are above 60.}
\label{tab:annotator-agreement}
\vspace{-4mm}
\end{table}

\subsection{Regular Updates}
Our dataset is highly sensitive to time since the ground truth is evolving along the world development. Therefore, we commit to updating the dataset regularly and researchers are strongly encouraged to stay tuned with our latest version for evaluation. And the datasets are mainly calibrated with information from Chinese Internet.

\subsection{Data Statistics}
Due to limitations in automatic query generation by GPT-4 and SeqGPT from the first stage, our dataset has low \textbf{retention rate} in which only 44.6\% synthetic data are accepted by human annotators. Among the accepted data, 53.1\% of them still need further modifications because of improper questions or wrong answers. For \textbf{question tags}, we have relatively balanced distributions between \textit{fast-changing} and \textit{slow-changing} questions with fewer \textit{never-changing} questions. For \textbf{answer types}, we have biased distributions as nearly 70\% of entities extracted from passages lie in ``person'' and ``group'' categories. This is because most of entities in first stage by automatic generation are ``person'' and ``group''. However, question tags and answer types could be changed or calibrated over time by re-annotation of the dataset. These distribution graphs and more details about our dataset are in Appendix~\ref{sec:dataset_distribution}.

\subsection{Evaluation}
As our dataset is constructed from Internet, our evaluation setup is based on \textbf{retrieval-augmented question answering} with search engines while the evaluation methods are \textit{answer rate} and \textit{F1-recall}. Our experiments consist of two settings: \textbf{close-book} and \textbf{open-book}. Overall, our evaluation provides a comprehensive understanding of current LLMs in factuality, especially for evolving knowledge. Besides, due to the safety implementation for different LLMs from helpful and harmless responses in training data \cite{bai2022training}, \textbf{F1-recall only counts on questions with effective responses} while \textbf{answer rate is used in representing the ratio of answered questions to the total questions}, which is a practical metric for the real world application of LLMs and could directly indicate the degree of hallucination in generated responses.

\paragraph{Evaluation Metrics} 
For \textbf{F1-recall}, we calculate \textit{the ratio of common tokens between model-generated responses and ground truth to the ground truth}. Specifically, we first segment the generated text and golden text into token lists using word segmentation tools~\footnote{\url{https://github.com/fxsjy/jieba}}, then calculate the ratio of tokens generated by models belonging to the golden token list to golden tokens.
For \textbf{answer rate}, we directly calculate \textit{the ratio of effectively answered questions to total questions}, i.e., responses of refusal are filtered out in our evaluation.

\section{Experiments}

\begin{figure}
  \center{
  \includegraphics
  [width=\linewidth]
  {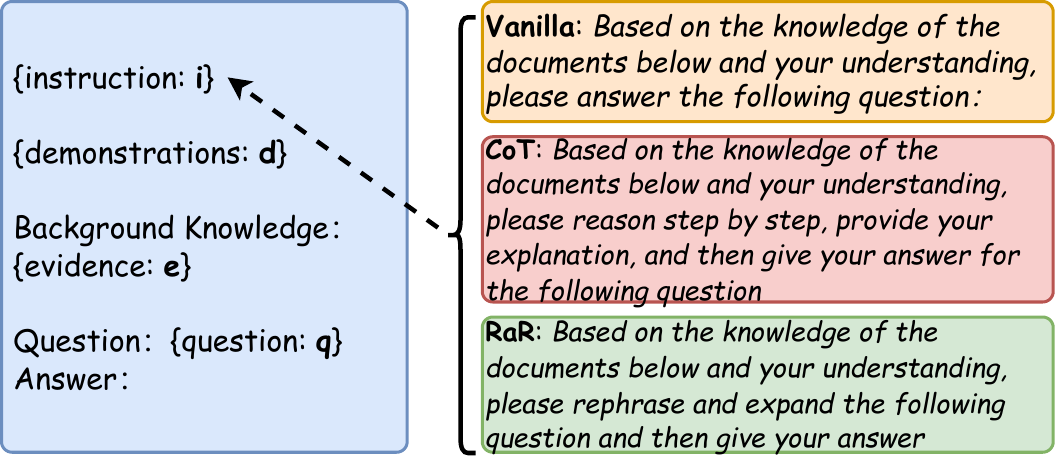}}
  \caption{Our prompts are formulated under this framework. Different prompting methods are used with different instructions $\mathbf{i}$. The Chinese version is in Appendix~\ref{sec:translated_prompt}.}
  \label{fig:prompt_framework}
\end{figure}

\begin{table*}[ht]
\tiny
\setlength\tabcolsep{0pt}
\centering\setcellgapes{1pt}\makegapedcells
\begin{tabular*}{\linewidth}{@{\extracolsep{\fill}}@{}lccccccccc@{}}
\toprule
\multicolumn{1}{c}{\multirow{2}{*}{\textbf{Model}}} &
  \multicolumn{3}{c}{\textbf{fast-changing}} &
  \multicolumn{3}{c}{\textbf{slow-changing}} &
  \multicolumn{3}{c}{\textbf{never-changing}} \\
\multicolumn{1}{c}{} &
  \textbf{+0} &
  \textbf{+5} &
  \textbf{+10} & 
  \textbf{+0} &
  \textbf{+5} &
  \textbf{+10} &
  \textbf{+0} &
  \textbf{+5} &
  \textbf{+10} \\ \cmidrule(l){2-4}\cmidrule(l){5-7}\cmidrule(l){8-10}
Intermlm-20B-Chat &
  16.1 [16] &
  49.7 [0] &
  50.5 [0] &
  19.8 [16] &
  61.2 [16] &
  61.8 [0] &
  34.0 [16] &
  68.1 [16] &
  71.7 [16] \\
Aquila2-34B-Chat &
  14.9 [16] &
  49.4 (99.6\%) [0] &
  51.5 [0] &
  17.7 [16] &
  60.4 [0] &
  62.5 [0] &
  35.6 [5] &
  69.2 [0] &
  69.4 [0] \\
Yi-34B-Chat &
  22.9 [16] &
  54.9 [16] &
  56.5 [16] &
  30.8 [16] &
  67.0 [16] &
  68.8 [16] &
  46.9 [16] &
  76.0 [16] &
  76.9 [16] \\
Deepseek-67B-Chat &
  24.3 [0] &
  57.5 [16] &
  58.4 [16] &
  \textbf{37.2 [16]} &
  \textbf{68.3 [16]} &
  \textbf{70.0 [16]} &
  53.1 [0] &
  \textbf{78.8 [16]} &
  \textbf{79.2 [5]} \\
ChatGPT &
  18.1 (96.6\%) [0] &
  56.2 (98.3\%) [0] &
  59.2 (98.3\%) [0] &
  14.1 (93.3\%) [0] &
  64.7 (97.9\%) [0] &
  66.3 (98.3\%) [0] &
  34.7 (99.0\%) [0] &
  73.2 [0] &
  73.7 (99.7\%) [0] \\
GPT-4 &
  \textbf{35.1 (13.5\%) [0]} &
  \textbf{59.6 (96.0\%) [0]} &
  \textbf{61.2 (96.4\%) [0]} &
  33.8 (25.4\%) [0] &
  65.8 (97.5\%) [0] &
  68.4 (96.5\%) [0] &
  \textbf{54.4 (56.1\%) [0]} &
  76.4 (99.3\%) [5] &
  78.8 (98.6\%) [5] \\
\bottomrule
\end{tabular*}
\caption{\label{tab:google_vanilla}Best performance over few-shot settings of baselines for \textbf{Vanilla} prompt with searched results from Google. +5 and +10 represent different numbers of searched results appended in the inputs. We report in the form of \textit{F1-recall (answer rate) [best number of few shot examples]} and omit the answer rate if it is 100\%. Data with the highest F1-recall scores are marked in bold.}
\end{table*}

\begin{table*}[ht]
\tiny
\setlength\tabcolsep{0pt}
\centering\setcellgapes{1pt}\makegapedcells
\begin{tabular*}{\linewidth}{@{\extracolsep{\fill}}@{}lccccccccc@{}}
\toprule
\multicolumn{1}{c}{\multirow{2}{*}{\textbf{Model}}} &
  \multicolumn{3}{c}{\textbf{fast-changing}} &
  \multicolumn{3}{c}{\textbf{slow-changing}} &
  \multicolumn{3}{c}{\textbf{never-changing}} \\
\multicolumn{1}{c}{} &
  \textbf{+0} &
  \textbf{+5} &
  \textbf{+10} &
  \textbf{+0} &
  \textbf{+5} &
  \textbf{+10} &
  \textbf{+0} &
  \textbf{+5} &
  \textbf{+10} \\ \cmidrule(l){2-4}\cmidrule(l){5-7}\cmidrule(l){8-10}
Intermlm-20B-Chat &
  15.0 [16] &
  46.8 [5] &
  49.5 [5] &
  18.1 [16] &
  59.3 [16] &
  61.7 [0] &
  34.7 [16] &
  67.2 [16] &
  68.9 [16] \\
Aquila2-34B-Chat &
  14.5 [16] &
  50.0 (99.6\%) [0] &
  51.9 [0] &
  17.1 [16] &
  60.0 [0] &
  61.4 [0] &
  35.6 [5] &
  69.5 [0] &
  69.8 [0] \\
Yi-34B-Chat &
  \textbf{23.2 [16]} &
  54.7 [5] &
  57.4 [16] &
  30.4 [16] &
  67.8 [16] &
  68.5 [16] &
  47.0 [16] &
  75.3 [16] &
  77.3 [16] \\
Deepseek-67B-Chat &
  22.9 [16] &
  58.0 [16] &
  59.2 [16] &
  \textbf{37.0 [5]} &
  67.8 [16] &
  70.6 [16] &
  \textbf{53.0 [16]} &
  79.8 [16] &
  80.2 [16] \\
ChatGPT &
  17.9 (97.3\%) [0] &
  57.6 (97.0\%) [0] &
  61.4 (96.6\%) [0] &
  13.9 (98.3\%) [0] &
  65.5 (97.5\%) [0] &
  65.7 (98.7\%) [0] &
  36.0 (99.7\%) [0] &
  75.2 (98.6\%) [0] &
  74.9 (98.6\%) [0] \\
GPT-4 &
  22.1 (82.9\%) [16] &
  \textbf{67.3 (89.0\%) [0]} &
  \textbf{68.0 (89.0\%) [5]} &
  19.8 (86.7\%) [0] &
  \textbf{73.9 (93.3\%) [0]} &
  \textbf{74.7 (90.4\%) [5]} &
  48.2 (56.1\%) [0] &
  \textbf{81.6 (98.3\%) [5]} &
  \textbf{83.5 (98.3\%) [0]} \\
\bottomrule
\end{tabular*}
\caption{\label{tab:google_cot}Best performance over few-shot settings of baselines for \textbf{CoT} prompt with searched results from Google. +5 and +10 represent different numbers of searched results appended in the inputs. We report in the form of \textit{F1-recall (answer rate) [best number of few shot examples]} and omit the answer rate if it is 100\%. Data with the highest F1-recall scores are marked in bold.}
\end{table*}

\begin{table*}[ht]
\tiny
\setlength\tabcolsep{0pt}
\centering\setcellgapes{1pt}\makegapedcells
\begin{tabular*}{\linewidth}{@{\extracolsep{\fill}}@{}lccccccccc@{}}
\toprule
\multicolumn{1}{c}{\multirow{2}{*}{\textbf{Model}}} &
  \multicolumn{3}{c}{\textbf{fast-changing}} &
  \multicolumn{3}{c}{\textbf{slow-changing}} &
  \multicolumn{3}{c}{\textbf{never-changing}} \\
\multicolumn{1}{c}{} & \textbf{+0} & \textbf{+5} & \textbf{+10} & \textbf{+0}      & \textbf{+5}     & \textbf{+10}    & \textbf{+0} & \textbf{+5}     & \textbf{+10}    \\ \cmidrule(l){2-4}\cmidrule(l){5-7}\cmidrule(l){8-10}
Intermlm-20B-Chat    & 16.9 [16]    & 49.5 [0]     & 52.4 [16]     & 20.0 [16]         & 61.2 [5]         & 63.3 [0]         & 34.6 [5]     & 68.2 [16]        & 71.3 [16]        \\
Aquila2-34B-Chat     & 15.5 [16]    & 48.4 [16]    & 51.4 [0]      & 17.5 [16]         & 59.6 [16]        & 61.9 [0]         & 36.1 [16]    & 68.0 [16]        & 69.5 [0]         \\
Yi-34B-Chat          & 22.8 [16]    & 54.6 [16]    & 57.0 [16]     & 30.6 [16]         & 68.2 [16]        & 68.5 [0]         & 47.7 [16]    & 77.1 [5]         & 76.8 [16]        \\
Deepseek-67B-Chat    & 23.3 [5]     & 57.6 [16]    & 58.9 [5]      & \textbf{37.7 [5]} & 68.5 [16]        & 70.7 [16]        & 54.2 [16]    & 79.6 [5]         & 79.8 [5]         \\
ChatGPT   & 19.2 [0]     & 58.9 [0]     & 61.7 [0]      & 15.9 [0]          & 65.8 (99.4\%) [0] & 67.6 (99.6\%) [0] & 35.6 [0]     & 75.2 (99.7\%) [0] & 76.5 (99.7\%) [0] \\
GPT-4 &
  \textbf{29.1 (37.3\%) [0]} &
  \textbf{61.8 (94.5\%) [0]} &
  \textbf{64.9 (95.2\%) [0]} &
  28.3 (54.62\%) [0] &
  \textbf{71.6 (94.2\%) [0]} &
  \textbf{71.6 (96.2\%) [0]} &
  \textbf{54.9 (72.5\%) [0]} &
  \textbf{80.7 (99.0\%) [0]} &
  \textbf{82.6 (99.3\%) [0]} \\
\bottomrule
\end{tabular*}
\caption{\label{tab:google_rar}Best performance over few-shot settings of baselines for \textbf{RaR} prompt with searched results from Google. +5 and +10 represent different numbers of searched results appended in the inputs. We report in the form of \textit{F1-recall(answer rate)[best number of few shot examples]} and omit the answer rate if it is 100\%. Data with the highest F1-recall scores are marked in bold.}
\end{table*}

\subsection{Experiment Setup} 
\paragraph{Baselines} We experiment with a series of baseline models pretrained with Chinese data, including representative OpenAI's \textbf{ChatGPT} (\textit{gpt-3.5-turbo-1106}) \cite{chatgpt} and \textbf{GPT-4} (\textit{gpt-4-1106-preview}) \cite{openai2023gpt4}, open-sourced Chinese-oriented models such as \textbf{Internlm-20B-Chat} \cite{2023internlm}, \textbf{Aquila2-34B-Chat} \cite{aquila2}, \textbf{Yi-34B-Chat} \cite{yi-model}, \textbf{Deepseek-67B-Chat} \cite{deepseekai2024deepseek}. In the close-book scenario, we only use the standalone LLM in answering questions with zero-shot and few-shot settings. For the open-book scenario, search engines are used for retrieving question-related results on the Internet and then fed into LLMs for reading.

\paragraph{Search Engines} Except for language models for information synthesis, we select three representative search engines to recall relevant passages from the Chinese Internet namely \textbf{Google} and \textbf{Bing}. These search engines are mainly used by Chinese people for information seeking. \textbf{Baidu} is omitted due to the difficulty in scraping its contents. The \textbf{Top-5} and \textbf{Top-10} searched results are provided to models in the \textbf{open-book setting}.

\paragraph{Prompt Design} Our prompt framework, which is in Chinese, could be framed as concatenation of $(\mathbf{i}, \mathbf{d}, \mathbf{e}, \mathbf{q})$, in Figure~\ref{fig:prompt_framework} where $\mathbf{i}$ represents the instruction, $\mathbf{d}$ for question-answer pairs from crowdsourced labelers, $\mathbf{e}$ for search results and $\mathbf{q}$ for current question. Different instructions \textbf{i} are used with three widely adopted prompting styles, \textbf{Vanilla}, \textbf{Chain-of-Thought (CoT)} \cite{wei2023chainofthought} and \textbf{Rephrase-and-Respond (RaR)} \cite{deng2023rephrase}. \textbf{Vanilla} instruction is directly asking models to answer questions with the context. \textbf{CoT} instruction is asking models to first explain and analyze the question $\mathbf{q}$ step by step and then give their answers. \textbf{RaR} instruction, however, is asking models to first rephrase and expand the question $\mathbf{q}$ and then give their answers, which could be viewed as a complement of CoT \cite{deng2023rephrase} as CoT is for diving deeper while RaR is for exploring broader. Besides, for demonstrations $\mathbf{d}$, we have used zero-shot and different few-shot settings, i.e., 5-shot and 16-shot. More specifically, our few-shot demonstrations are made up of human written questions and answers similar to \textit{CDQA} dataset without contexts or other explanations as it costs longer time without any improvement.

\subsection{Results and Analyses}
Table~\ref{tab:google_vanilla},~\ref{tab:google_cot},~\ref{tab:google_rar} summarize best performances over few-shot prompting across different baselines for Vanilla, CoT and RaR prompts respectively. Our default search engine for analysis is \textbf{Google} as it is most widely used around the world.


\paragraph{Baseline Comparison} In Table~\ref{tab:google_cot}, we see that \textbf{GPT-4} achieves the best across all questions. Only GPT-4 reaches nearly or over 65, 70 and 80 in F1-recall for \textit{fast-changing}, \textit{slow-changing} and \textit{never-changing} questions respectively with searched results which indicates that there is still room of improvement for other open-sourced models with Chinese instructions under the retrieval-augmented-generation scenario. Besides, the second-best model, \textbf{Deepseek-67B-Chat}, has shown great performance as it even surpasses GPT-4 on \textit{slow-changing} and \textit{never-changing} questions by Vanilla prompt with searched results in Table~\ref{tab:google_vanilla} and performs better (both in answer rate and F1-recall) than GPT-4 without searched results in \textit{slow-changing} and \textit{never-changing} questions especially in CoT prompt. This indicates that Deepseek-67B-Chat has stored more Chinese Internet knowledge as its alignment corpora mainly focus on Chinese and English while GPT-4 aligns with multilingual data. Moreover, it is worth noting that answer rates of ChatGPT and GPT-4 are often lower than 100\% across all different prompts and question types especially for GPT-4 in close-book question answering which indicates that \textbf{there are more hallucination reduction measures such as refusal of questions in GPT-4 than other models}.

\begin{figure}[ht]
    \begin{subfigure}{\linewidth}
        \centering
        \includegraphics[width=1.0\linewidth, center]{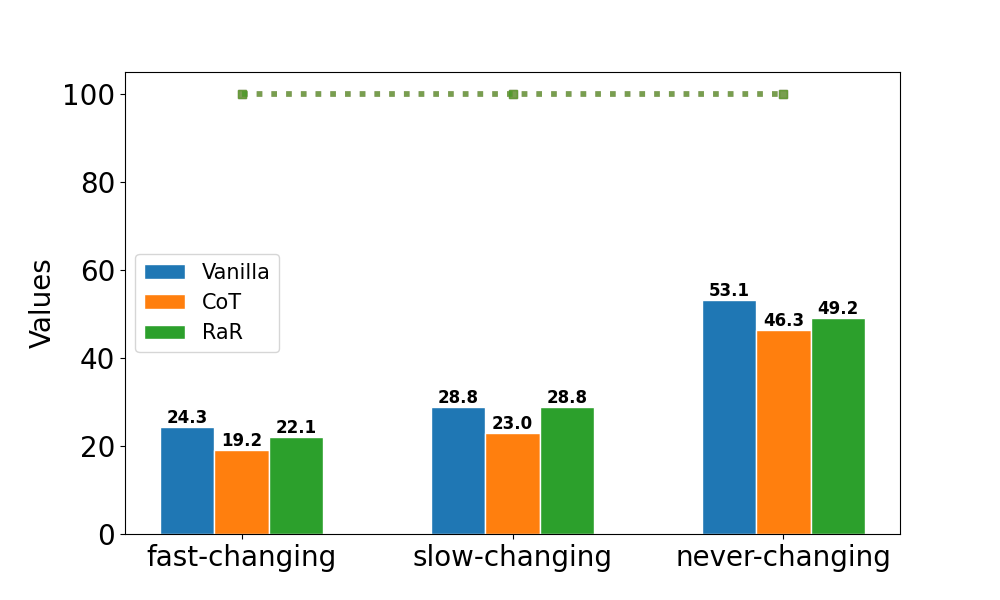}
        \caption{\label{fig:close-book-compare-prompts-deepseek}Deepseek-67B-Chat}
    \end{subfigure}
    \hfill
    \begin{subfigure}{\linewidth}
        \centering
        \includegraphics[width=1.0\linewidth, center]{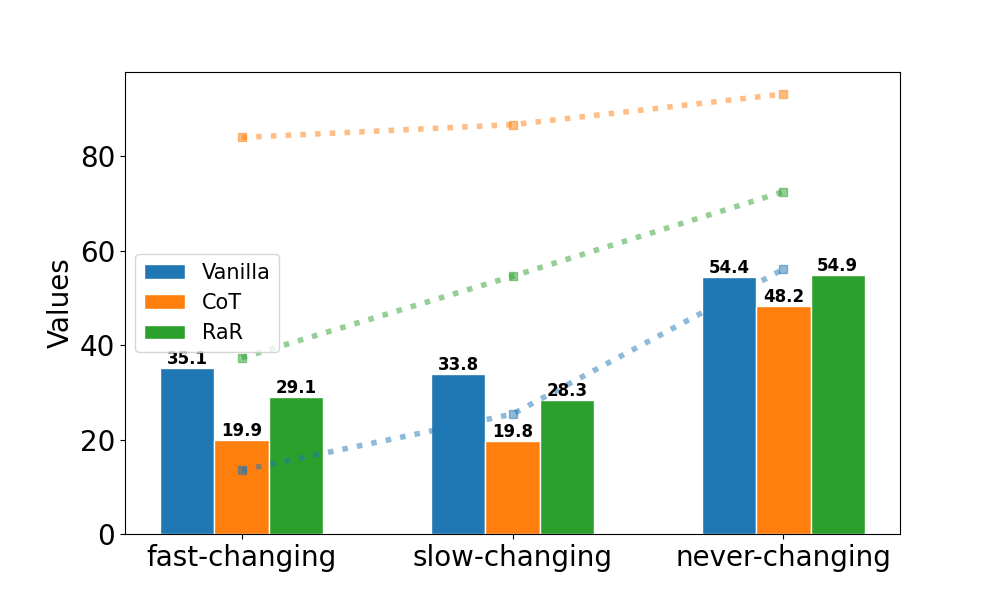}
        \caption{\label{fig:close-book-compare-prompts-gpt4}GPT-4}
    \end{subfigure}
  \caption{\label{fig:close-book-compare-prompts}F1-recall scores and Answer Rates of \textbf{different prompts} for LLMs in \textbf{close-book} scenario under zero-shot setting. We represent F1-recall scores with bar plots and answer rates with dotted lines.}
\end{figure}

\paragraph{How do different styles of prompts work in LLMs?} As our evaluation scenarios are made up of close-book and open-book, different prompts just elicit different behaviors in LLMs. To rule out the influence of few-shot demonstrations, we use zero-shot setting in this analysis. We use GPT-4 and Deepseek models as closed and open-sourced representative models in the following analysis. In Figure~\ref{fig:close-book-compare-prompts}, under \textbf{close-book} scenario, we see that Deepseek-67B-Chat has same answer rates over different prompts on different questions while there are different answering behaviors in GPT-4. Specifically, GPT-4 answers with great care in vanilla prompts with lowest answer rates but high F1-recall scores while GPT-4 suffers from hallucination in CoT and RaR prompts with at most +522\% and +176\% in answer rates but -43\% and -17\% in F1-recall scores compared to Vanilla prompt. For both models, Vanilla prompts outperform the other two kinds of prompts. \textbf{This indicates that verbose explanation or expansion could increase hallucination especially when without evidence}.

\begin{figure}[ht]
    \begin{subfigure}{\linewidth}
        \centering
        \includegraphics[width=1.0\linewidth, center]{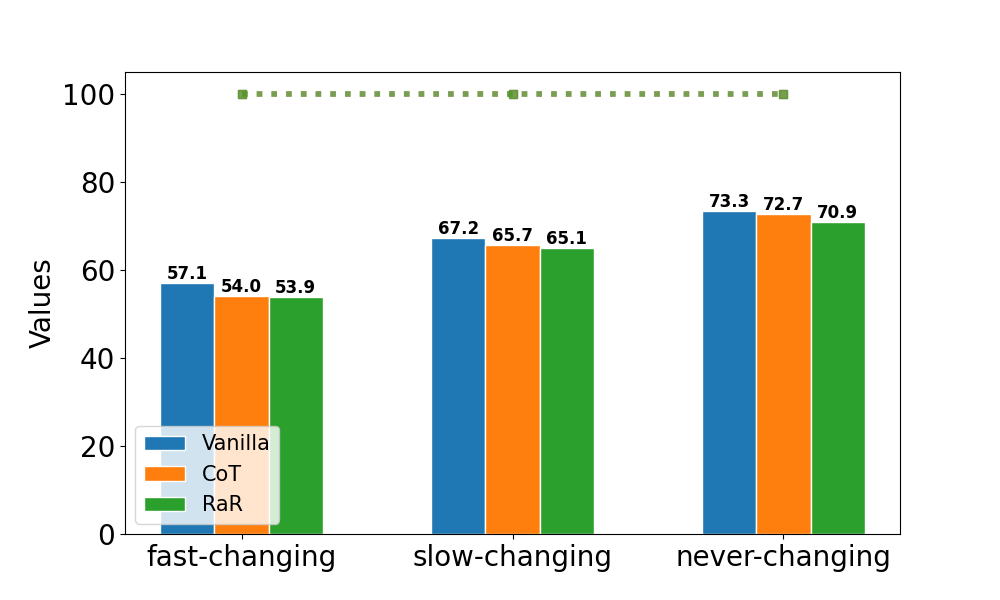}
        \caption{\label{fig:open-book-compare-prompts-deepseek}Deepseek-67B-Chat}
    \end{subfigure}
    \hfill
    \begin{subfigure}{\linewidth}
        \centering
        \includegraphics[width=1.0\linewidth, center]{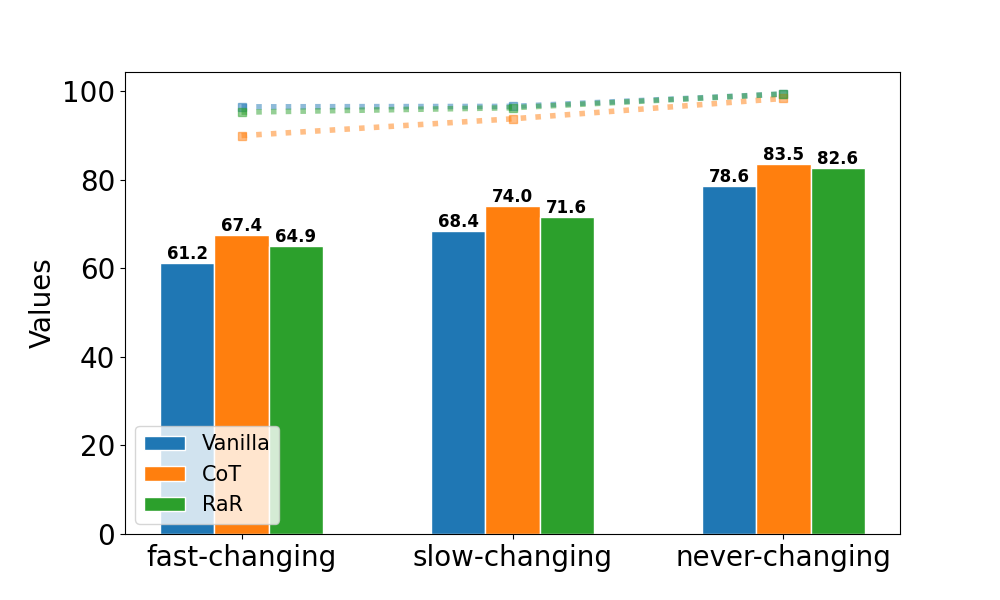}
        \caption{\label{fig:open-book-compare-prompts-gpt4}GPT-4}
    \end{subfigure}
  \caption{\label{fig:open-book-compare-prompts}F1-recall scores and Answer Rates of \textbf{different prompts} for LLMs in \textbf{open-book} scenario under zero-shot setting. We represent F1-recall scores with bar plots and answer rates with dotted lines.}
\end{figure}

In Figure~\ref{fig:open-book-compare-prompts}, under \textbf{open-book} scenario\footnote{We use Top-10 searched results as more searched results generally improve the results of LLMs from Table~\ref{tab:google_vanilla},~\ref{tab:google_cot},~\ref{tab:google_rar}.}, we see that Deepseek-67B-Chat still shares same answer rates across different prompts and question types and vanilla prompt is the best for it, which indicates that CoT and RaR take risks in inducing more hallucinated answers. For GPT-4, answer rates have less gaps between different prompts (all over 90\%) and F1-recall scores are all increasing dramatically compared to close-book counterparts, representing adding contextual information elicits LLMs in answering questions more efficiently. Besides, with search results, CoT and RaR both outperform Vanilla prompt and CoT performs the best in GPT-4 with less hallucination, i.e., lower answer rate, and higher F1-recall score. \textbf{This indicates that CoT and RaR could improve LLMs on complex tasks but CoT elicits more reasoning abilities to directly improve the answering}.

\begin{figure}[ht]
    \centering
    \includegraphics[width=1.0\linewidth, center]{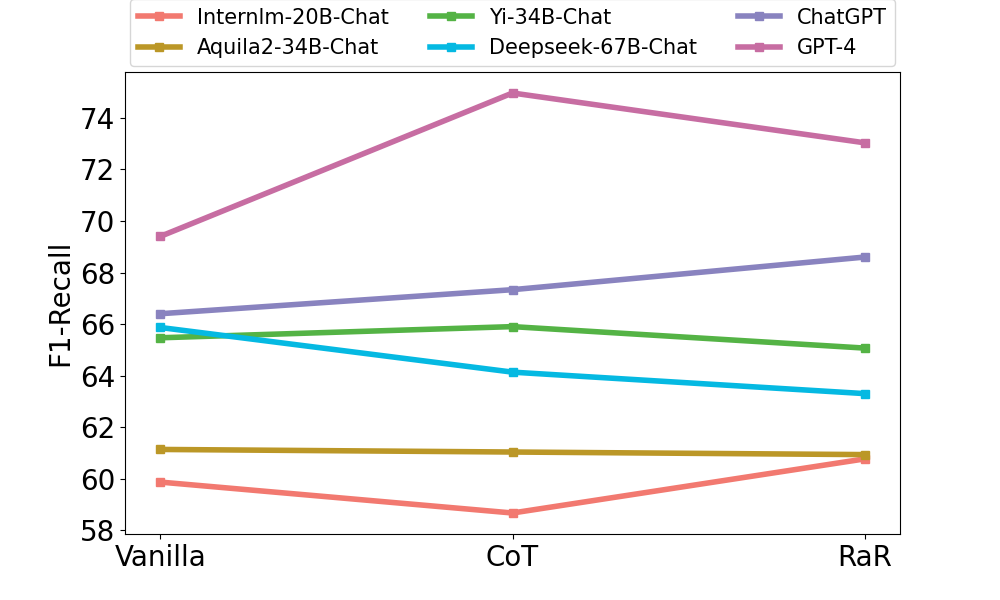}
    \caption{\label{fig:open-book-model-size-prompts-compare}F1-recall scores averaged over all three different questions for all models with \textbf{different prompts} in open-book scenario under zero-shot setting. We present F1-recall score only since all answer rates $\geq$ 90\%.}
\end{figure}

Nevertheless, model sizes and training data are both fundamental for these prompts to work. In Figure~\ref{fig:open-book-model-size-prompts-compare}, \textbf{not every model improves with CoT or RaR compared to Vanilla prompt}. For example, Deepseek-34B-Chat's performance decrease in CoT and RaR; ChatGPT and Internlm-20B-Chat prefer RaR while GPT-4 and Yi-34B-Chat prefer CoT for larger gains; Aquila2-34B-Chat is robust to all prompt types.

\begin{figure}[ht]
    \begin{subfigure}{\linewidth}
        \centering
        \includegraphics[width=1.0\linewidth, center]{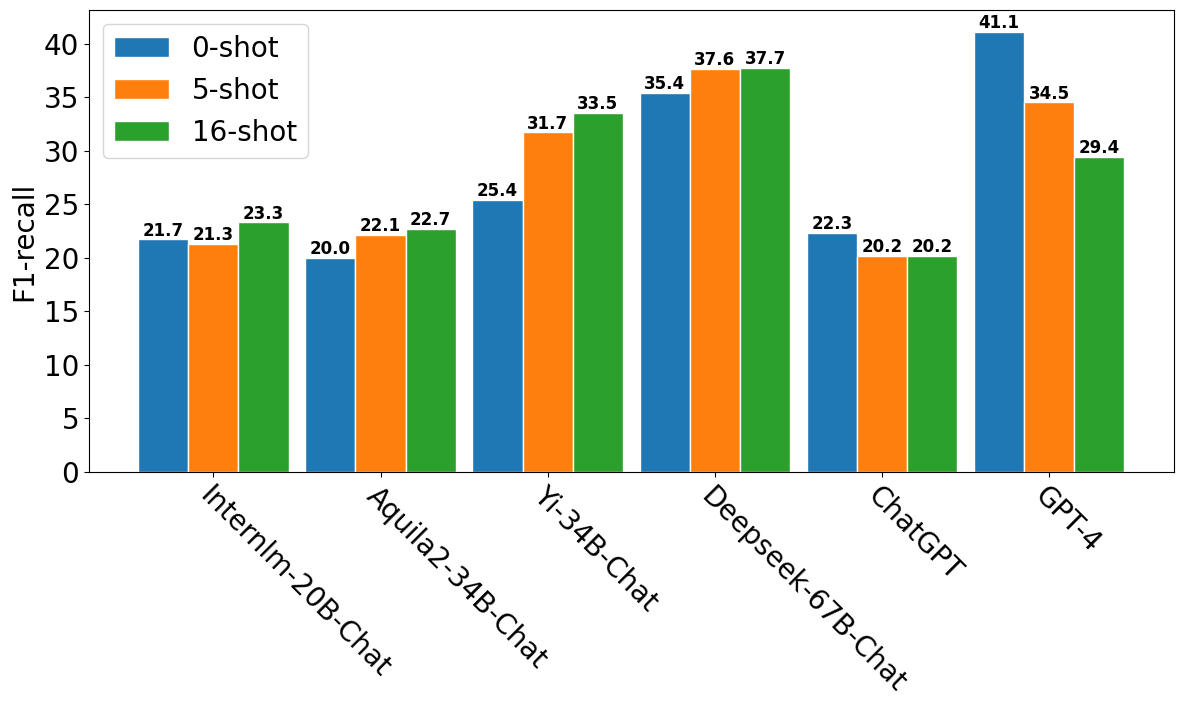}
        \caption{\label{fig:close-book-few-shot-compare}Close-book}
    \end{subfigure}
    \hfill
    \begin{subfigure}{\linewidth}
        \centering
        \includegraphics[width=1.0\linewidth, center]{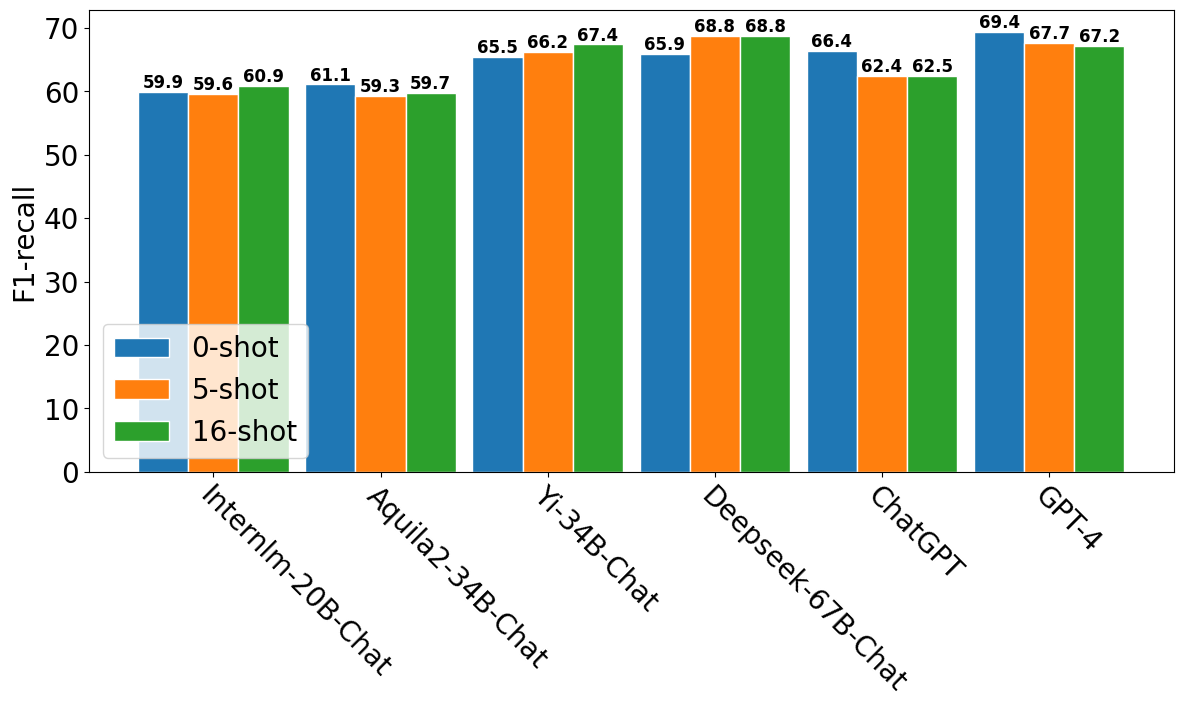}
        \caption{\label{fig:open-book-few-shot-compare}Open-book (Top-10 searched results)}
    \end{subfigure}
  \caption{\label{fig:few-shot-compare}F1-recall scores averaged over all types of questions for different models with \textbf{different few-shot settings} in close-book and open-book scenarios.}
\end{figure}

\paragraph{Does few-shot prompting always work for all LLMs?} For better analyzing the influence of few-shot prompting, we collect close-book and open-book results of all LLMs in \textbf{zero-shot} setting and \textbf{vanilla} prompt. In Figure~\ref{fig:few-shot-compare}, based on nearly 100\% answer rate, five in close-book scenario and four in open-book scenario out of all five open-sourced models have better performance with more few-shot demonstrations, which are sampled in the same data distribution during generating \textit{CDQA}.

\begin{table}[ht]
\footnotesize
\setlength\tabcolsep{0pt}
\centering\setcellgapes{6pt}\makegapedcells
\begin{tabular*}{\linewidth}{@{\extracolsep{\fill}}@{}lcccccc@{}}
\toprule
\multicolumn{1}{c}{\multirow{2}{*}{\textbf{Models}}} & \multicolumn{3}{c}{\textbf{close-book}} & \multicolumn{3}{c}{\textbf{open-book}} \\ \cmidrule(l){2-4}\cmidrule(l){5-7} 
\multicolumn{1}{c}{} & 0-shot & 5-shot & 16-shot & 0-shot & 5-shot & 16-shot \\
ChatGPT              & 96.3   & 95.0   & 96.7    & 98.8   & 99.7   & \textbf{99.9}    \\
GPT-4                & 31.7   & 52.2   & 64.7    & 97.4   & 97.7   & \textbf{98.0}    \\ \bottomrule
\end{tabular*}
\caption{\label{tab:gpt-answer-rates}Answer rates for ChatGPT and GPT-4 averaged on all types of questions in close-book and open-book scenarios with \textbf{different few-shot settings}.}
\end{table}
However, we also notice that ChatGPT and GPT-4 have shown different trends compared to open-sourced models, i.e., more few-shot examples lead to decreases in F1-recall scores. Therefore, we check the answer rates of ChatGPT and GPT-4 in Table~\ref{tab:gpt-answer-rates} where ChatGPT stays in fairly high answer rates ($\geq 95\%$) and GPT-4 increases its answer rates with more few-shot examples. Combined with their monotonic decrease in F1-recall scores, we show that ChatGPT and GPT-4 hallucinate more with more few-shot examples in \textit{CDQA}. \textbf{This indicates the challenge of \textit{CDQA} and different LLMs with different capabilities struggle in different ways}.

\begin{figure}[ht]
    \centering
    \includegraphics[width=0.95\linewidth, center]{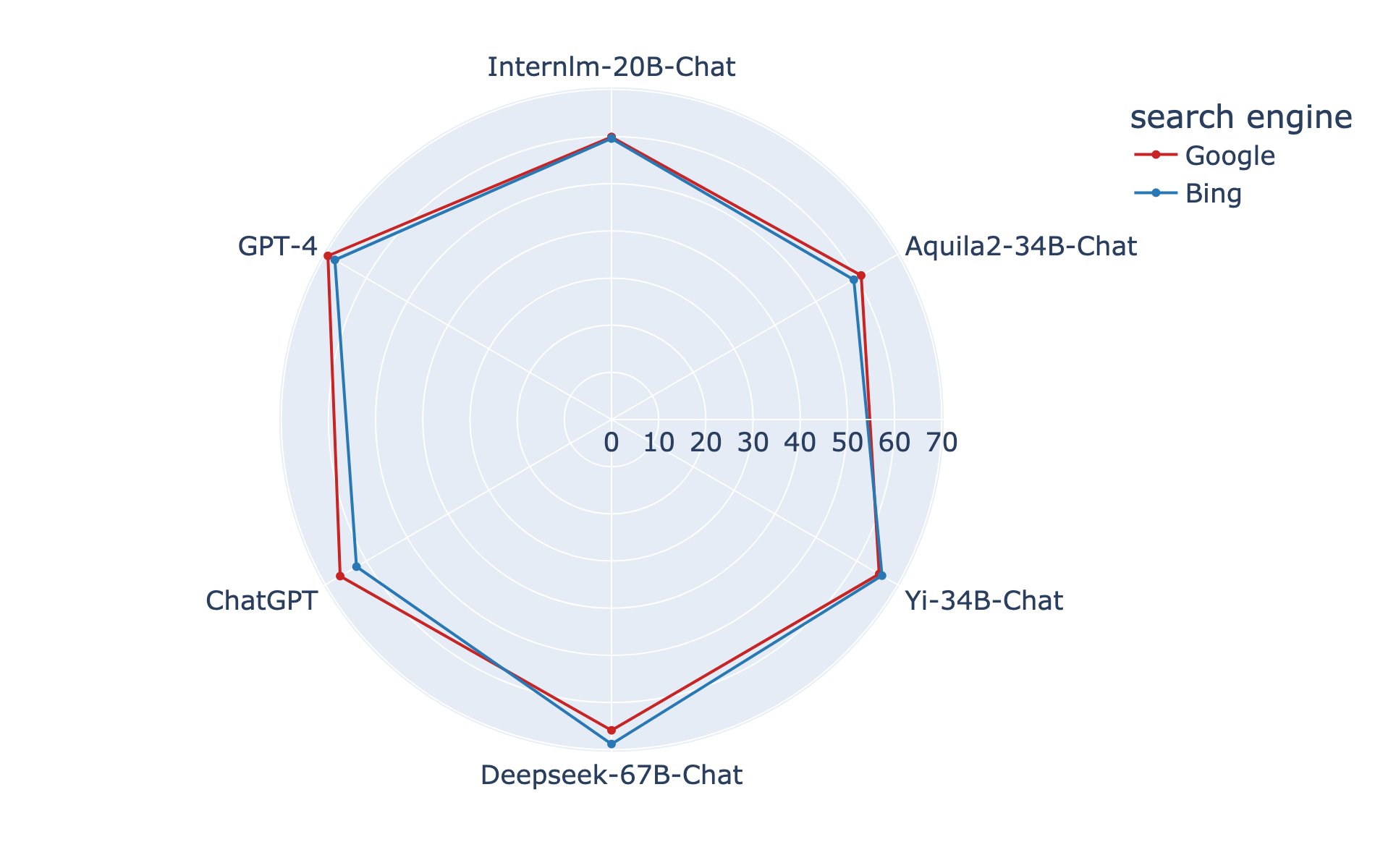}
    \caption{\label{fig:full-search-engine-compare}F1-recall scores averaged over all questions for different models with \textbf{different search engines}.}
\end{figure}

\paragraph{How do different search engines help?} For a more detailed comparison between search engines across all baselines, we use vanilla prompt since CoT performs better in LLMs and zero-shot setting with Top-10 searched results from different search engines. 
In Figure~\ref{fig:full-search-engine-compare}, searched results from Google slightly outperforms Bing among all baseline models, which indicates that the \textbf{Google currently provides the most useful retrieved evidence for question answering about Chinese news}. 

\begin{figure}[ht]
    \centering
    \includegraphics[width=0.95\linewidth, center]{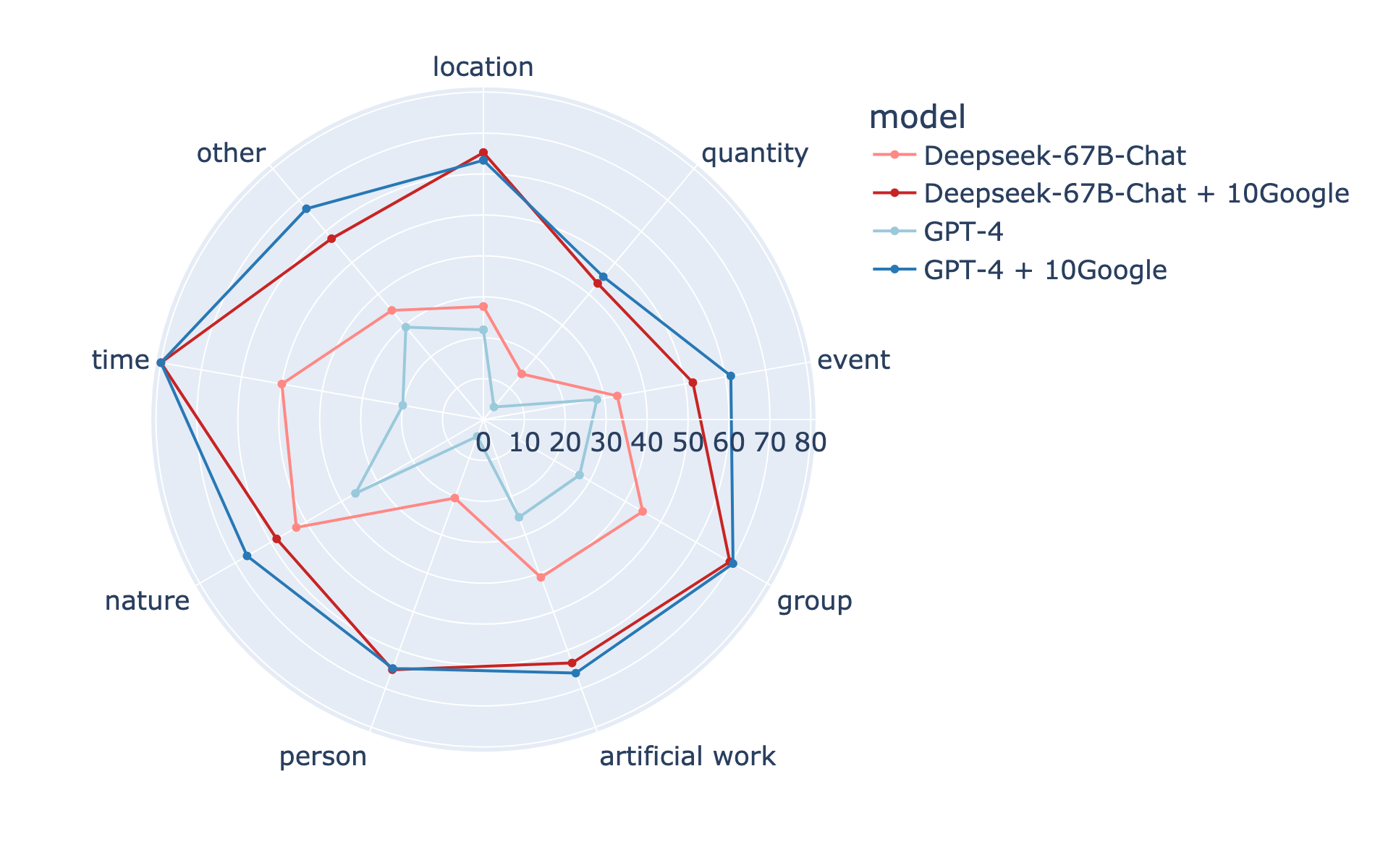}
    \caption{\label{fig:answer-types-compare}F1-recall scores on \textbf{different answer types} for Deepseek-67B-Chat and GPT-4 in close-book and open-book scenarios with Vanilla prompt. We use Top 10 searched results from Google. Deepseek-67B-Chat holds 100\% answer rate in all time while GPT-4 drastically increase its answer rate to 100\% with searched results from Google.}
\end{figure}

\paragraph{How do LLMs perform across different answer types?} As answers in \textit{CDQA} are mainly entities from news, we conduct analysis across different answer types for two representative open-sourced and closed LLMs, i.e., Deepseek-67B-Chat and GPT-4. In Figure~\ref{fig:answer-types-compare}, we observe that \textbf{GPT-4's internal knowledge is poorer than Chinese-oriented models such as Deepseek-67B-Chat for Chinese users}. However, with enough retrieved evidence, \textbf{GPT-4 has stronger abilities in learning from contexts than Deepseek-67B-Chat} where this ``learning efficiency'', i.e., \textit{the ratio of gaps between open-book scores and close-book scores to the close-book} could reach at most 1370\% compared to 219\% in Deepseek-67B-Chat.
Moreover, from Figure~\ref{fig:answer-types-compare}, we also could notice that the ``quantity'' group is hardest for models to figure out the correct answers, which is due to the granularity of answers and the need of reasoning abilities.

\section{Related Work} 

\paragraph{Temporal and Dynamic QA Benchmark} \textit{StreamingQA} \cite{liska2022streamingqa} is a QA dataset where questions are human writen or LM-generated on given dates, showing how open-book and close-book QA models adapt to new knowledge over time and importance of retrieval augmentation in up-to-date search space. \textit{TimeQA} \cite{chen2021dataset} are formed from extracted evolving facts in \textit{WikiData} by manual extraction and verification while we extract entities to directly formulate them as answer candidates based on the documents. \textit{RealTimeQA} \cite{kasai2022realtime}, a dynamic QA benchmark with automatic weekly updates from the weekly News Quiz section in social media such as CNN, is most related to our semi-automatic question generation with latest Chinese news corpus.

\paragraph{English QA benchmark}
Question answering is a long-standing task in NLP area~\cite{wang2024evaluating, li2024llms}, ranging from classic single-turn benchmarks such as \textit{SQuAD} \cite{rajpurkar2016squad, rajpurkar2018know}, \textit{TriviaQA} \cite{joshi2017triviaqa} and \textit{Natural Questions} \cite{naturalquestions} to conversational QA like \textit{TopiOCQA} \cite{adlakha2022topiocqa}. Also, it could also be categorized by normal human intelligence and human experts intelligence. For example, \textit{CommonsenseQA} \cite{talmor2019commonsenseqa} comes up with questions from \textit{ConceptNet} in discriminating the target concepts with commonsense while \textit{GPQA} \cite{rein2023gpqa} consists of questions generated from graduate-level students. Although these benchmarks have provided efficient and effective evaluation metrics and covered a variety of topics, they are mostly static with a focus on less-evolving knowledge, which might already be or going to be included in the pre-training data of large language models.

\paragraph{Chinese QA benchmark} In contrast to prosperous English QA benchmarks, Chinese counterparts are still under-explored. \textit{DuReader} \cite{he2018dureader} is a classic free-form QA benchmark collected by Baidu from its own products and \textit{CLUE} \cite{xu2020clue} is the first large scale NLU benchmark in Chinese. After the recent debut of powerful large language models, a series of Chinese QA benchmarks are proposed for better evaluating them. \textit{C-Eval} \cite{huang2023ceval} is a multiple-choice questions answering dataset from Chinese Standard Exams. \textit{WebCPM} \cite{qin2023webcpm} collects questions from web forums through web searching and browsing and \textit{SuperCLUE} \cite{xu2023superclue} is a comprehensive Chinese benchmark for question answering in aligning users needs. But they all suffer from either data leakage or the risk of saturated performance which hinders the accurate evaluation on questions requiring fresh knowledge to answer as static questions are readily overfitted.


\section{Conclusion}
The creation of \textit{CDQA} addresses the urgent need for the evaluation of Chinese LLMs, thereby improving LLM-driven applications for Chinese users. Given the cultural influences in LLMs' training data, it is our aspiration that \textit{CDQA} will foster development in various capabilities of LLMs, particularly within Chinese contexts. While \textit{CDQA} progresses further with a semi-automatic generation pipeline with more data than \textit{FreshQA}, we acknowledge that it is far from a perfect LLM evaluation. Other critical dimensions, including tool learning, LLMs safety, and robustness, remain to be explored. 
However, we believe that our constructed \textit{CDQA} and the series of insights obtained based on it will provide valuable resources and guidance for subsequent research on Chinese LLMs.
In the future, we will conduct more in-depth analyses of the capabilities of LLMs based on \textit{CDQA} and investigate how to enhance the LLMs’ ability to handle dynamic questions. This will empower LLMs to better cope with the complex and ever-changing real-world application environments.

\clearpage
\section*{Limitations}
One of the limitations of our work is that the language we study is Chinese only. As the two most widely used languages in the world, English and Chinese have always been equally valued and widely concerned in the NLP community. In fact, our work is inspired by previous \textit{FreshQA} in the English scenario and aims to provide similar data resources to Chinese LLMs researchers. We also encourage and welcome more researchers from other languages to engage in similar research.

In addition, another limitation that cannot be ignored is how to keep our \textit{CDQA} updated. Because \textit{CDQA} focuses on questions whose answers change dynamically, it is critical to ensure that the answers to questions in \textit{CDQA} are always correct and up-to-date. Therefore, we also commit to updating our \textit{CDQA} regularly and providing researchers with the latest version of \textit{CDQA} for LLMs evaluation. 

\section*{Ethics Statement}
The task we focus on is the evaluation of LLMs, and the LLMs we evaluate are all public and widely used LLMs, so they do not bring potential ethical risks.
The data samples of \textit{CDQA} that we collect have been manually cleaned and pre-processed to ensure that they do not contain any data that will cause moral risks, such as politically sensitive, violent, and private data.
In addition, we also have signed legal labor contracts with the human annotators we employ, and pay them higher than market prices based on their workload.


\bibliography{anthology,custom}

\begin{thebibliography}{36}
\expandafter\ifx\csname natexlab\endcsname\relax\def\natexlab#1{#1}\fi

\bibitem[{01-ai(2023)}]{yi-model}
01-ai. 2023.
\newblock Yi series large langugae models.
\newblock \url{https://github.com/01-ai/Yi}.

\bibitem[{Adlakha et~al.(2022)Adlakha, Dhuliawala, Suleman, de~Vries, and Reddy}]{adlakha2022topiocqa}
Vaibhav Adlakha, Shehzaad Dhuliawala, Kaheer Suleman, Harm de~Vries, and Siva Reddy. 2022.
\newblock \href {http://arxiv.org/abs/2110.00768} {Topiocqa: Open-domain conversational question answering with topic switching}.

\bibitem[{BAAI(2023)}]{aquila2}
BAAI. 2023.
\newblock Aquila2.
\newblock \url{https://github.com/FlagAI-Open/Aquila2}.

\bibitem[{Bai et~al.(2022)Bai, Jones, Ndousse, Askell, Chen, DasSarma, Drain, Fort, Ganguli, Henighan, Joseph, Kadavath, Kernion, Conerly, El-Showk, Elhage, Hatfield-Dodds, Hernandez, Hume, Johnston, Kravec, Lovitt, Nanda, Olsson, Amodei, Brown, Clark, McCandlish, Olah, Mann, and Kaplan}]{bai2022training}
Yuntao Bai, Andy Jones, Kamal Ndousse, Amanda Askell, Anna Chen, Nova DasSarma, Dawn Drain, Stanislav Fort, Deep Ganguli, Tom Henighan, Nicholas Joseph, Saurav Kadavath, Jackson Kernion, Tom Conerly, Sheer El-Showk, Nelson Elhage, Zac Hatfield-Dodds, Danny Hernandez, Tristan Hume, Scott Johnston, Shauna Kravec, Liane Lovitt, Neel Nanda, Catherine Olsson, Dario Amodei, Tom Brown, Jack Clark, Sam McCandlish, Chris Olah, Ben Mann, and Jared Kaplan. 2022.
\newblock \href {http://arxiv.org/abs/2204.05862} {Training a helpful and harmless assistant with reinforcement learning from human feedback}.

\bibitem[{Bosselut et~al.(2022)Bosselut, Li, Lin, Shwartz, Majumder, Lal, Rudinger, Ren, Tandon, and Zouhar}]{csrr-2022-commonsense}
Antoine Bosselut, Xiang Li, Bill~Yuchen Lin, Vered Shwartz, Bodhisattwa~Prasad Majumder, Yash~Kumar Lal, Rachel Rudinger, Xiang Ren, Niket Tandon, and Vil{\'e}m Zouhar, editors. 2022.
\newblock \href {https://aclanthology.org/2022.csrr-1.0} {\emph{Proceedings of the First Workshop on Commonsense Representation and Reasoning (CSRR 2022)}}. Association for Computational Linguistics, Dublin, Ireland.

\bibitem[{Chang et~al.(2023)Chang, Wang, Wang, Wu, Yang, Zhu, Chen, Yi, Wang, Wang et~al.}]{chang2023survey}
Yupeng Chang, Xu~Wang, Jindong Wang, Yuan Wu, Linyi Yang, Kaijie Zhu, Hao Chen, Xiaoyuan Yi, Cunxiang Wang, Yidong Wang, et~al. 2023.
\newblock A survey on evaluation of large language models.
\newblock \emph{ACM Transactions on Intelligent Systems and Technology}.

\bibitem[{Chen et~al.(2021)Chen, Wang, and Wang}]{chen2021dataset}
Wenhu Chen, Xinyi Wang, and William~Yang Wang. 2021.
\newblock \href {http://arxiv.org/abs/2108.06314} {A dataset for answering time-sensitive questions}.

\bibitem[{DeepSeek-AI(2024)}]{deepseekai2024deepseek}
DeepSeek-AI. 2024.
\newblock \href {http://arxiv.org/abs/2401.02954} {Deepseek llm: Scaling open-source language models with longtermism}.

\bibitem[{Deng et~al.(2023)Deng, Zhang, Chen, and Gu}]{deng2023rephrase}
Yihe Deng, Weitong Zhang, Zixiang Chen, and Quanquan Gu. 2023.
\newblock \href {http://arxiv.org/abs/2311.04205} {Rephrase and respond: Let large language models ask better questions for themselves}.

\bibitem[{He et~al.(2018)He, Liu, Liu, Lyu, Zhao, Xiao, Liu, Wang, Wu, She, Liu, Wu, and Wang}]{he2018dureader}
Wei He, Kai Liu, Jing Liu, Yajuan Lyu, Shiqi Zhao, Xinyan Xiao, Yuan Liu, Yizhong Wang, Hua Wu, Qiaoqiao She, Xuan Liu, Tian Wu, and Haifeng Wang. 2018.
\newblock \href {http://arxiv.org/abs/1711.05073} {Dureader: a chinese machine reading comprehension dataset from real-world applications}.

\bibitem[{Huang et~al.(2023{\natexlab{a}})Huang, Ma, Li, Huang, Zou, Zhang, and Zheng}]{huang2023lateval}
Shulin Huang, Shirong Ma, Yinghui Li, Mengzuo Huang, Wuhe Zou, Weidong Zhang, and Hai-Tao Zheng. 2023{\natexlab{a}}.
\newblock Lateval: An interactive llms evaluation benchmark with incomplete information from lateral thinking puzzles.
\newblock \emph{arXiv preprint arXiv:2308.10855}.

\bibitem[{Huang et~al.(2023{\natexlab{b}})Huang, Bai, Zhu, Zhang, Zhang, Su, Liu, Lv, Zhang, Lei, Fu, Sun, and He}]{huang2023ceval}
Yuzhen Huang, Yuzhuo Bai, Zhihao Zhu, Junlei Zhang, Jinghan Zhang, Tangjun Su, Junteng Liu, Chuancheng Lv, Yikai Zhang, Jiayi Lei, Yao Fu, Maosong Sun, and Junxian He. 2023{\natexlab{b}}.
\newblock \href {http://arxiv.org/abs/2305.08322} {C-eval: A multi-level multi-discipline chinese evaluation suite for foundation models}.

\bibitem[{InternLMTeam(2023)}]{2023internlm}
InternLMTeam. 2023.
\newblock Internlm: A multilingual language model with progressively enhanced capabilities.
\newblock \url{https://github.com/InternLM/InternLM}.

\bibitem[{Joshi et~al.(2017)Joshi, Choi, Weld, and Zettlemoyer}]{joshi2017triviaqa}
Mandar Joshi, Eunsol Choi, Daniel~S. Weld, and Luke Zettlemoyer. 2017.
\newblock \href {http://arxiv.org/abs/1705.03551} {Triviaqa: A large scale distantly supervised challenge dataset for reading comprehension}.

\bibitem[{Kasai et~al.(2022)Kasai, Sakaguchi, Takahashi, Bras, Asai, Yu, Radev, Smith, Choi, and Inui}]{kasai2022realtime}
Jungo Kasai, Keisuke Sakaguchi, Yoichi Takahashi, Ronan~Le Bras, Akari Asai, Xinyan Yu, Dragomir Radev, Noah~A. Smith, Yejin Choi, and Kentaro Inui. 2022.
\newblock \href {http://arxiv.org/abs/2207.13332} {Realtime qa: What's the answer right now?}

\bibitem[{Kwiatkowski et~al.(2019)Kwiatkowski, Palomaki, Redfield, Collins, Parikh, Alberti, Epstein, Polosukhin, Kelcey, Devlin, Lee, Toutanova, Jones, Chang, Dai, Uszkoreit, Le, and Petrov}]{naturalquestions}
Tom Kwiatkowski, Jennimaria Palomaki, Olivia Redfield, Michael Collins, Ankur Parikh, Chris Alberti, Danielle Epstein, Illia Polosukhin, Matthew Kelcey, Jacob Devlin, Kenton Lee, Kristina~N. Toutanova, Llion Jones, Ming-Wei Chang, Andrew Dai, Jakob Uszkoreit, Quoc Le, and Slav Petrov. 2019.
\newblock Natural questions: a benchmark for question answering research.
\newblock \emph{Transactions of the Association of Computational Linguistics}.

\bibitem[{Li et~al.(2023)Li, Huang, Ma, Jiang, Li, Zhou, Zheng, and Zhou}]{DBLP:journals/corr/abs-2307-09007}
Yinghui Li, Haojing Huang, Shirong Ma, Yong Jiang, Yangning Li, Feng Zhou, Hai{-}Tao Zheng, and Qingyu Zhou. 2023.
\newblock \href {https://doi.org/10.48550/ARXIV.2307.09007} {On the (in)effectiveness of large language models for chinese text correction}.
\newblock \emph{CoRR}, abs/2307.09007.

\bibitem[{Li et~al.(2024)Li, Zhou, Luo, Ma, Li, Zheng, Hu, and Yu}]{li2024llms}
Yinghui Li, Qingyu Zhou, Yuanzhen Luo, Shirong Ma, Yangning Li, Hai-Tao Zheng, Xuming Hu, and Philip~S Yu. 2024.
\newblock When llms meet cunning questions: A fallacy understanding benchmark for large language models.
\newblock \emph{arXiv preprint arXiv:2402.11100}.

\bibitem[{Liska et~al.(2022)Liska, Kocisky, Gribovskaya, Terzi, Sezener, Agrawal, Cyprien De~Masson, Scholtes, Zaheer, Young et~al.}]{liska2022streamingqa}
Adam Liska, Tomas Kocisky, Elena Gribovskaya, Tayfun Terzi, Eren Sezener, Devang Agrawal, D’Autume Cyprien De~Masson, Tim Scholtes, Manzil Zaheer, Susannah Young, et~al. 2022.
\newblock Streamingqa: A benchmark for adaptation to new knowledge over time in question answering models.
\newblock In \emph{International Conference on Machine Learning}, pages 13604--13622. PMLR.

\bibitem[{McHugh(2012)}]{cohenkappa}
Mary McHugh. 2012.
\newblock \href {https://doi.org/10.11613/BM.2012.031} {Interrater reliability: The kappa statistic}.
\newblock \emph{Biochemia medica : časopis Hrvatskoga društva medicinskih biokemičara / HDMB}, 22:276--82.

\bibitem[{OpenAI(2022)}]{chatgpt}
OpenAI. 2022.
\newblock Chatgpt.
\newblock \url{https://chat.openai.com/chat}.

\bibitem[{OpenAI(2023)}]{openai2023gpt4}
OpenAI. 2023.
\newblock \href {http://arxiv.org/abs/2303.08774} {Gpt-4 technical report}.

\bibitem[{Pan et~al.(2024)Pan, Luo, Wang, Chen, Wang, and Wu}]{pan2024unifying}
Shirui Pan, Linhao Luo, Yufei Wang, Chen Chen, Jiapu Wang, and Xindong Wu. 2024.
\newblock Unifying large language models and knowledge graphs: A roadmap.
\newblock \emph{IEEE Transactions on Knowledge and Data Engineering}.

\bibitem[{Qin et~al.(2023)Qin, Cai, Jin, Yan, Liang, Zhu, Lin, Han, Ding, Wang, Xie, Qi, Liu, Sun, and Zhou}]{qin2023webcpm}
Yujia Qin, Zihan Cai, Dian Jin, Lan Yan, Shihao Liang, Kunlun Zhu, Yankai Lin, Xu~Han, Ning Ding, Huadong Wang, Ruobing Xie, Fanchao Qi, Zhiyuan Liu, Maosong Sun, and Jie Zhou. 2023.
\newblock \href {http://arxiv.org/abs/2305.06849} {Webcpm: Interactive web search for chinese long-form question answering}.

\bibitem[{Rajpurkar et~al.(2018)Rajpurkar, Jia, and Liang}]{rajpurkar2018know}
Pranav Rajpurkar, Robin Jia, and Percy Liang. 2018.
\newblock \href {http://arxiv.org/abs/1806.03822} {Know what you don't know: Unanswerable questions for squad}.

\bibitem[{Rajpurkar et~al.(2016)Rajpurkar, Zhang, Lopyrev, and Liang}]{rajpurkar2016squad}
Pranav Rajpurkar, Jian Zhang, Konstantin Lopyrev, and Percy Liang. 2016.
\newblock \href {http://arxiv.org/abs/1606.05250} {Squad: 100,000+ questions for machine comprehension of text}.

\bibitem[{Rein et~al.(2023)Rein, Hou, Stickland, Petty, Pang, Dirani, Michael, and Bowman}]{rein2023gpqa}
David Rein, Betty~Li Hou, Asa~Cooper Stickland, Jackson Petty, Richard~Yuanzhe Pang, Julien Dirani, Julian Michael, and Samuel~R. Bowman. 2023.
\newblock \href {http://arxiv.org/abs/2311.12022} {Gpqa: A graduate-level google-proof q\&a benchmark}.

\bibitem[{Shanahan(2024)}]{shanahan2024talking}
Murray Shanahan. 2024.
\newblock Talking about large language models.
\newblock \emph{Communications of the ACM}, 67(2):68--79.

\bibitem[{Talmor et~al.(2019)Talmor, Herzig, Lourie, and Berant}]{talmor2019commonsenseqa}
Alon Talmor, Jonathan Herzig, Nicholas Lourie, and Jonathan Berant. 2019.
\newblock \href {http://arxiv.org/abs/1811.00937} {Commonsenseqa: A question answering challenge targeting commonsense knowledge}.

\bibitem[{Vu et~al.(2023)Vu, Iyyer, Wang, Constant, Wei, Wei, Tar, Sung, Zhou, Le, and Luong}]{vu2023freshllms}
Tu~Vu, Mohit Iyyer, Xuezhi Wang, Noah Constant, Jerry Wei, Jason Wei, Chris Tar, Yun-Hsuan Sung, Denny Zhou, Quoc Le, and Thang Luong. 2023.
\newblock \href {http://arxiv.org/abs/2310.03214} {Freshllms: Refreshing large language models with search engine augmentation}.

\bibitem[{Wang et~al.(2024)Wang, Cheng, Guo, Yue, Ding, Xu, Wang, Hu, Zhang, and Zhang}]{wang2024evaluating}
Cunxiang Wang, Sirui Cheng, Qipeng Guo, Yuanhao Yue, Bowen Ding, Zhikun Xu, Yidong Wang, Xiangkun Hu, Zheng Zhang, and Yue Zhang. 2024.
\newblock Evaluating open-qa evaluation.
\newblock \emph{Advances in Neural Information Processing Systems}, 36.

\bibitem[{Wei et~al.(2022)Wei, Tay, Bommasani, Raffel, Zoph, Borgeaud, Yogatama, Bosma, Zhou, Metzler et~al.}]{wei2022emergent}
Jason Wei, Yi~Tay, Rishi Bommasani, Colin Raffel, Barret Zoph, Sebastian Borgeaud, Dani Yogatama, Maarten Bosma, Denny Zhou, Donald Metzler, et~al. 2022.
\newblock Emergent abilities of large language models.
\newblock \emph{arXiv preprint arXiv:2206.07682}.

\bibitem[{Wei et~al.(2023)Wei, Wang, Schuurmans, Bosma, Ichter, Xia, Chi, Le, and Zhou}]{wei2023chainofthought}
Jason Wei, Xuezhi Wang, Dale Schuurmans, Maarten Bosma, Brian Ichter, Fei Xia, Ed~Chi, Quoc Le, and Denny Zhou. 2023.
\newblock \href {http://arxiv.org/abs/2201.11903} {Chain-of-thought prompting elicits reasoning in large language models}.

\bibitem[{Xu et~al.(2020)Xu, Hu, Zhang, Li, Cao, Li, Xu, Sun, Yu, Yu, Tian, Dong, Liu, Shi, Cui, Li, Zeng, Wang, Xie, Li, Patterson, Tian, Zhang, Zhou, Liu, Zhao, Zhao, Yue, Zhang, Yang, Richardson, and Lan}]{xu2020clue}
Liang Xu, Hai Hu, Xuanwei Zhang, Lu~Li, Chenjie Cao, Yudong Li, Yechen Xu, Kai Sun, Dian Yu, Cong Yu, Yin Tian, Qianqian Dong, Weitang Liu, Bo~Shi, Yiming Cui, Junyi Li, Jun Zeng, Rongzhao Wang, Weijian Xie, Yanting Li, Yina Patterson, Zuoyu Tian, Yiwen Zhang, He~Zhou, Shaoweihua Liu, Zhe Zhao, Qipeng Zhao, Cong Yue, Xinrui Zhang, Zhengliang Yang, Kyle Richardson, and Zhenzhong Lan. 2020.
\newblock \href {http://arxiv.org/abs/2004.05986} {Clue: A chinese language understanding evaluation benchmark}.

\bibitem[{Xu et~al.(2023)Xu, Li, Zhu, Xue, Zhu, Zhao, He, Zhang, Kang, and Lan}]{xu2023superclue}
Liang Xu, Anqi Li, Lei Zhu, Hang Xue, Changtai Zhu, Kangkang Zhao, Haonan He, Xuanwei Zhang, Qiyue Kang, and Zhenzhong Lan. 2023.
\newblock \href {http://arxiv.org/abs/2307.15020} {Superclue: A comprehensive chinese large language model benchmark}.

\bibitem[{Yu et~al.(2023)Yu, Jiang, Lou, Huang, Wang, Liu, Cai, Li, Li, Tu, Zheng, Zhang, Xie, Huang, and Jiang}]{yu2023seqgpt}
Tianyu Yu, Chengyue Jiang, Chao Lou, Shen Huang, Xiaobin Wang, Wei Liu, Jiong Cai, Yangning Li, Yinghui Li, Kewei Tu, Hai-Tao Zheng, Ningyu Zhang, Pengjun Xie, Fei Huang, and Yong Jiang. 2023.
\newblock \href {http://arxiv.org/abs/2308.10529} {Seqgpt: An out-of-the-box large language model for open domain sequence understanding}.

\end{thebibliography}
\bibliographystyle{acl_natbib}

\appendix

\section{Tag Taxonomy of CDQA}
\label{sec:tag_taxonomy}
The tag taxonomy of \textit{CDQA} is presented in Table~\ref{tab:tags_desc}.

\begin{table*}[htp]
\begin{tabularx}{\linewidth}{lXX}
\hline
\textbf{Category} & \textbf{Description} &\textbf{Example}\\
\hline
\rowcolor{c1}\texttt{fast-changing} & The answer to the question is prone to changing within \textbf{one year} & (\textit{How many sessions has the Maritime Silk Road Cultural Heritage Forum been held?, Four})\\
\rowcolor{c2}\texttt{slow-changing} & The answer to the question is prone to changing in \textbf{several years} & (\textit{Which ancient city site in China has recently been recognized as a UNESCO World Cultural Heritage?; the Liangzhu Ancient City Site})\\
\rowcolor{c3}\texttt{never-changing} & The answer to the question is from \textbf{static knowledge} such as scientific theories, historical facts and so on & (\textit{In rural areas during winter heating, it is necessary to guard against the risk of poisoning from which gas?; Carbon monoxide})\\
\Xhline{3\arrayrulewidth}
\rowcolor{c4}\texttt{person} & Specific individual, usually referring to a human being. & (\textit{Who among the current representatives of the Fuxin County People's Congress was one of the first batch of anti-epidemic heroes to rush to support Wuhan?; Xin Li})\\
\rowcolor{c5}\texttt{location} & Geographical position. & (\textit{Which province has recently strengthened the regulation of the intellectual property agency industry?; Hainan})\\
\rowcolor{c6}\texttt{time} & Points or intervals of a continuous sequence of events or conditions. & (\textit{In which year was the recent "Haikou Cup" sailing competition held?; 2023})\\
\rowcolor{c7}\texttt{event} & Something that happens, which can be planned or spontaneous. & (\textit{What themed event was recently launched in Suzhou High-speed Railway New Town to promote the development of private enterprises?; "Suzhou Sentiments, Private Enterprises Connected at Heart"})\\
\rowcolor{c8}\texttt{artificial work} & Items or intellectual achievements created by humans, which have artistic, academic, or practical value. & (\textit{What is the latest TV series aired starring Xin Jiang?; As Long As We Are Together})\\
\rowcolor{c9}\texttt{group} & Entities formed by multiple individuals for a specific purpose. & (\textit{Which undergraduate university is rencently established in the Ningxia Hui Autonomous Region recently?; Ningxia Minjiang Institute of Applied Technology})\\
\rowcolor{c10}\texttt{nature} & Phenomena or entities in the natural world. & (\textit{Please explain to me what Nucleases is?; Small RNA molecules with catalytic function, belonging to the category of biological catalysts?; capable of degrading specific mRNA sequences.})\\
\rowcolor{c11}\texttt{quantity} & Numeric value for times or stuff. & (\textit{How many base pairs in human Y chromosome have been observed from the latest sequencing results?；More than 30 million})\\
\rowcolor{c12}\texttt{other} & Other answer not classified to the above categories. & (\textit{Is there any fee for withdrawing WeChat balance to bank card?; Yes})\\
\hline
\end{tabularx}
\caption{\label{tab:tags_desc}Descriptions and examples of \textbf{question tags} (first three rows) and \textbf{answer types} (last nine rows). We represent (\textit{<question>}; \textit{<answer>}) as examples. Original language for these examples is Chinese. We translate them here for better preview.}
\end{table*}


\section{Dataset Distributions}
\label{sec:dataset_distribution}
\begin{figure}
  \center{
  \includegraphics
  [width=1.2\linewidth, center]
  {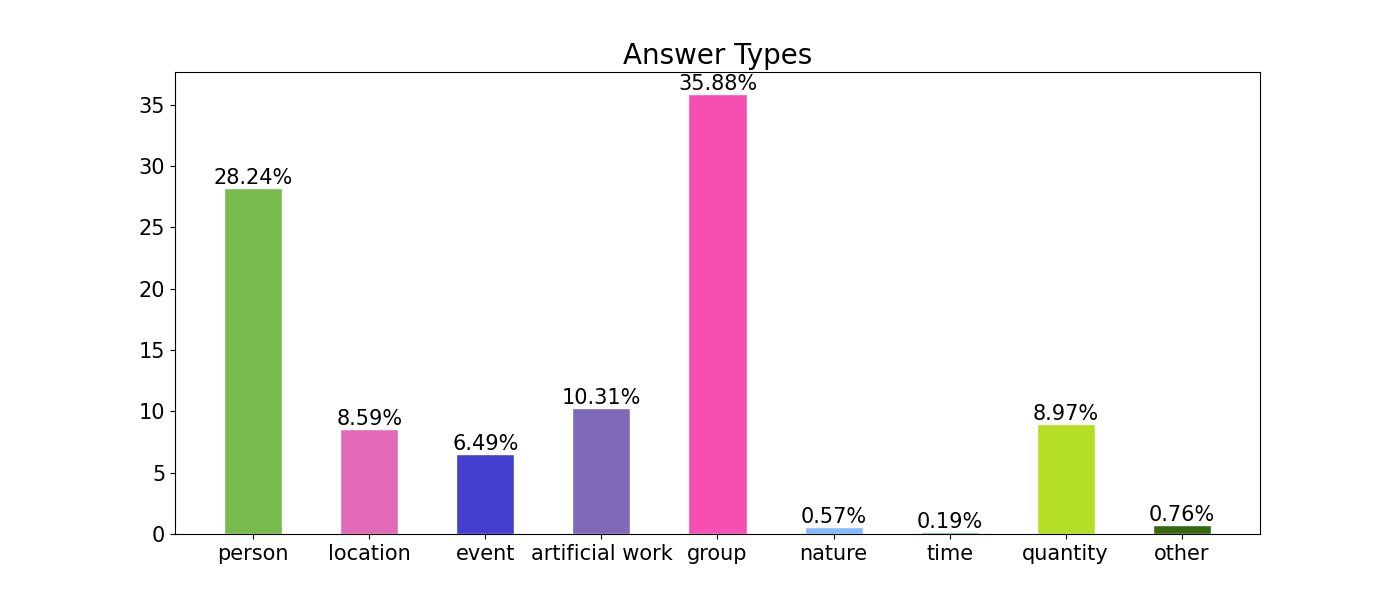}}
  \caption{\label{fig:answer_tag_fast}Distributions of answer types for fast-changing questions.}
\end{figure}

\begin{figure}
  \center{
  \includegraphics
  [width=1.2\linewidth, center]
  {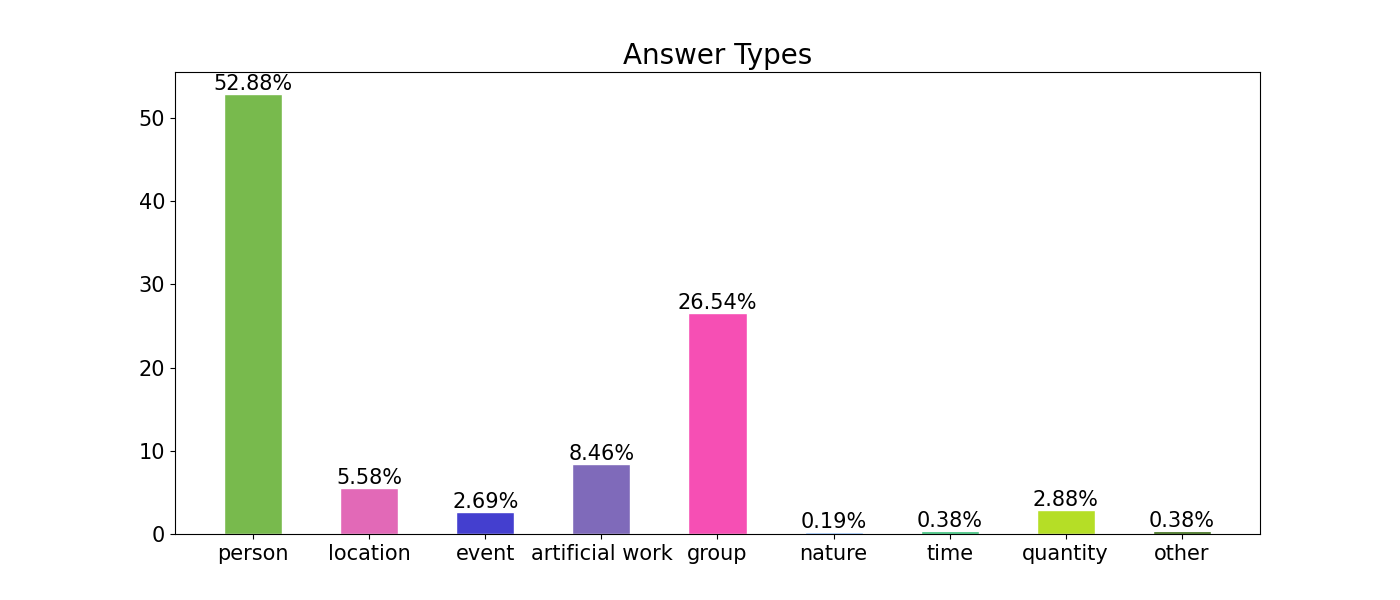}}
  \caption{\label{fig:answer_tag_slow}Distributions of answer types for slow-changing questions.}
\end{figure}

\begin{figure}
  \center{
  \includegraphics
  [width=1.2\linewidth, center]
  {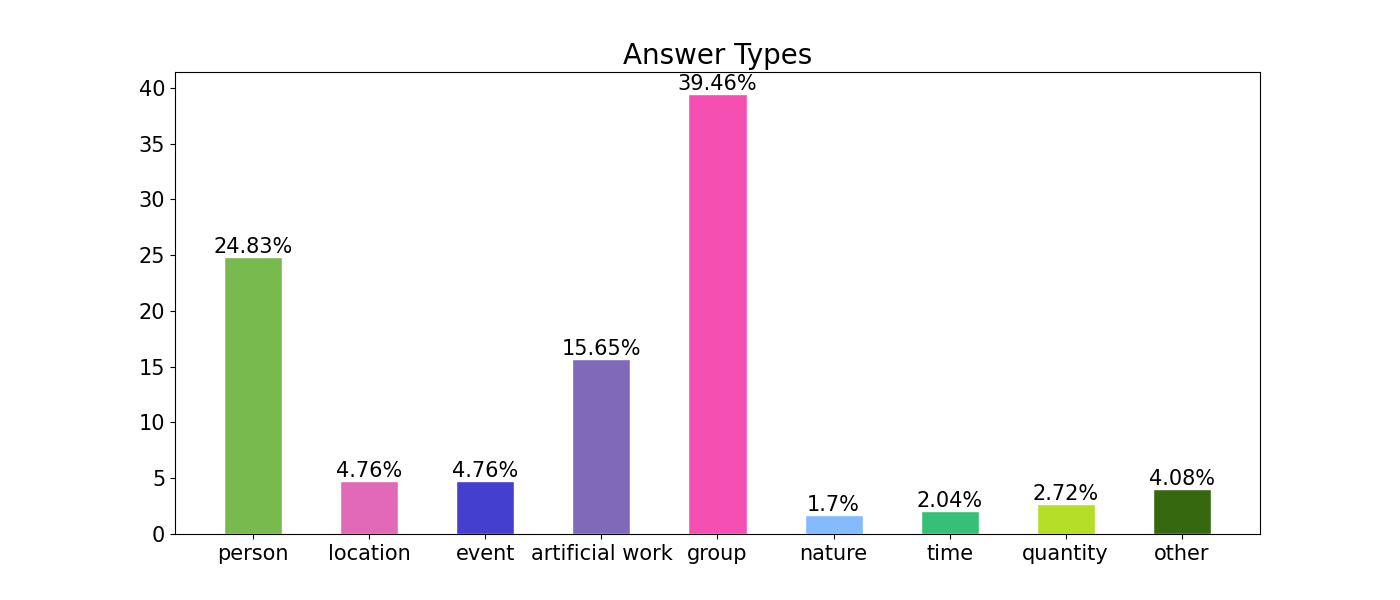}}
  \caption{\label{fig:answer_tag_never}Distributions of answer types for never-changing questions.}
\end{figure}

Knowledge types for queries and answer types are visualized in the following Figure~\ref{fig:question_tag},~\ref{fig:answer_type}. More specifically, we have further visualized the answer type distributions in each question tag. From Figure~\ref{fig:answer_tag_fast}, Figure~\ref{fig:answer_tag_slow} and Figure~\ref{fig:answer_tag_never}, we see that nearly 80\% of slow changing questions are about person and group. Except for person and group, artificial work should be the third largest category for answers, which includes jobs, titles, knowledge and so on. These observations are all consistent with our data sources as information for the protagonists, places and events are compulsory and most frequent in news reports. Besides, percentages of time reach the maximum in never-changing tag as currently most of questions answered with time are about the frequencies.
\begin{figure}
  \center{
  \includegraphics
  [width=1.2\linewidth, center]
  {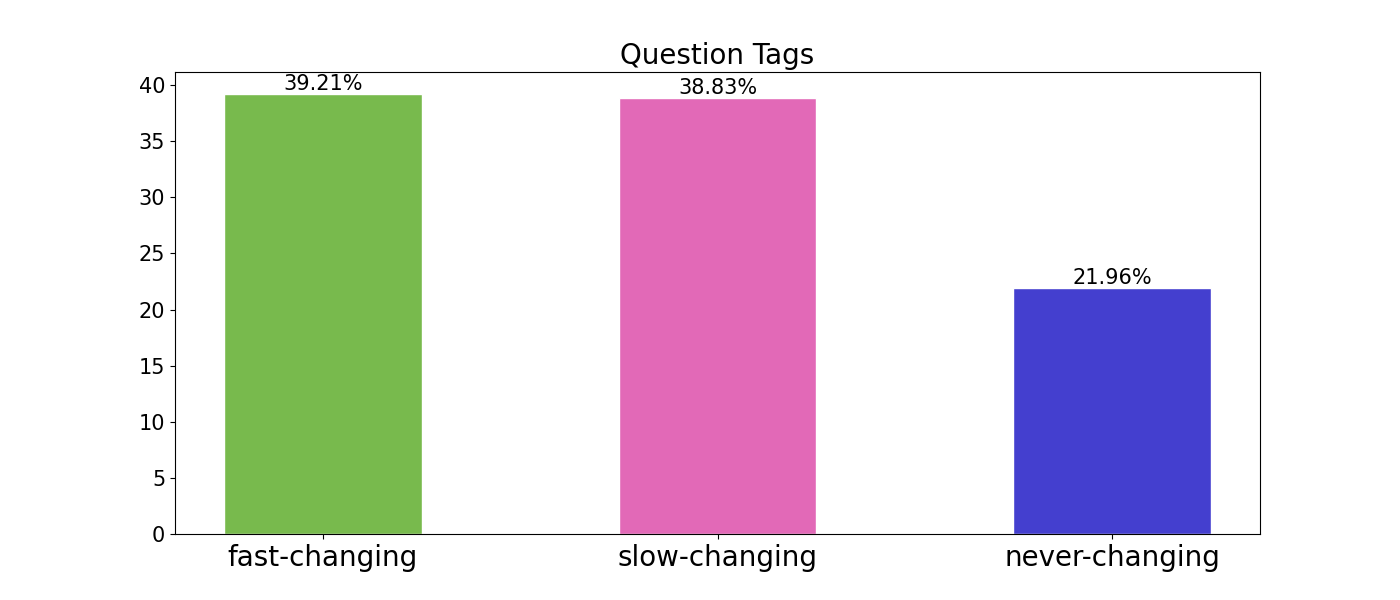}}
  \caption{\label{fig:question_tag}Distributions of question tags for full data.}
\end{figure}
\begin{figure}
  \center{
  \includegraphics
  [width=1.2\linewidth, center]
  {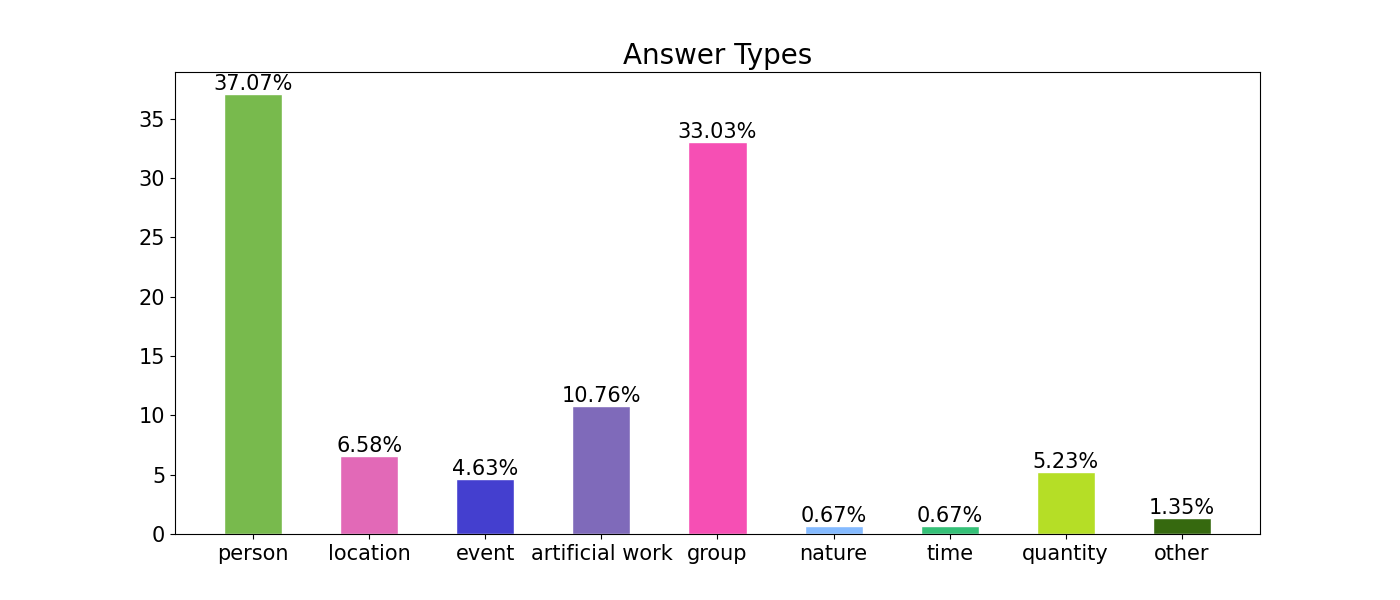}}
  \caption{\label{fig:answer_type}Distribution of answer types for full data.}
\end{figure}

\section{Translated Chinese Prompts}
\label{sec:translated_prompt}
\begin{figure}
  \center{
  \includegraphics
  [width=\linewidth]
  {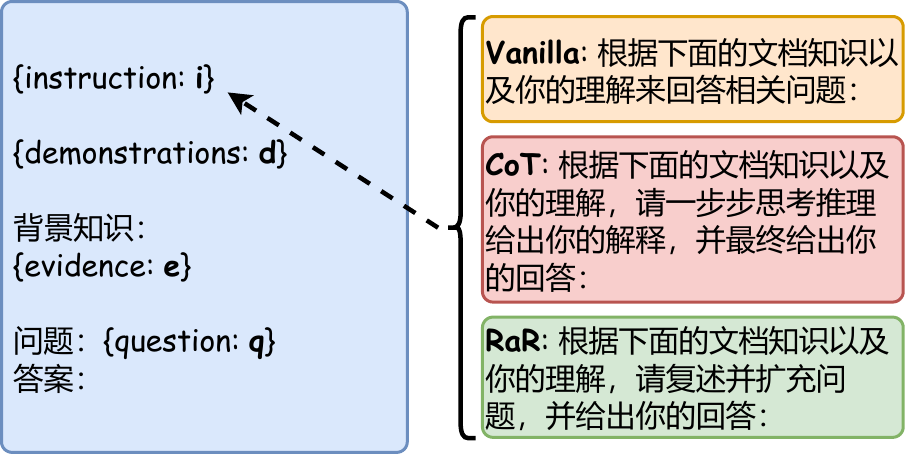}}
  \caption{The Chinese prompt framework for Figure~\ref{fig:prompt_framework}.}
  \label{fig:prompt_framework_chn}
\end{figure}

The translated prompt framework is illustrated in Figure~\ref{fig:prompt_framework_chn}.

\end{CJK*}

\end{document}